\documentclass{sn-jnl}

\usepackage{graphicx}%
\usepackage{multirow}%
\usepackage{amsmath,amssymb,amsfonts}%
\usepackage{mathtools}
\usepackage{amsthm}%
\usepackage{mathrsfs}%
\usepackage[title]{appendix}%
\usepackage{xcolor}%
\usepackage{textcomp}%
\usepackage{manyfoot}%
\usepackage{booktabs}%
\usepackage{algorithm}%
\usepackage{algorithmicx}%
\usepackage{algpseudocode}%
\usepackage{listings}
\usepackage{adjustbox}
\usepackage{gensymb}
\usepackage{rotating}
\usepackage[backend=biber,style=numeric,sorting=none]{biblatex}
\begin{document}



\title[Article Title]{Diffusion Bridge Networks Simulate Clinical-grade PET from MRI for Dementia Diagnostics}


\author*[1,2]{\fnm{Yitong} \sur{Li}}\email{yi\_tong@tum.de}

\author[3]{\fnm{Ralph} \sur{Buchert}}\email{r.buchert@uke.de}

\author[4]{\fnm{Benita} \sur{Schmitz-Koep}}\email{benita.schmitz-koep@tum.de}

\author[5]{\fnm{Timo} \sur{Grimmer}}\email{t.grimmer@tum.de}

\author[2,6]{\fnm{Björn} \sur{Ommer}}\email{b.ommer@lmu.de}

\author*[4]{\fnm{Dennis M.} \sur{Hedderich}}\email{dennis.hedderich@tum.de}
\equalcont{Equally contributed.}

\author*[7]{\fnm{Igor} \sur{Yakushev}}\email{igor.yakushev@tum.de}
\equalcont{Equally contributed.}

\author*[1,2]{\fnm{Christian} \sur{Wachinger}}\email{christian.wachinger@tum.de}
\equalcont{Equally contributed.}

\affil[1]{\orgdiv{Lab for AI in Medical Imaging, Institute for Diagnostic and Interventional Radiology, School of Medicine and Health}, \orgname{Technical University of Munich (TUM)}, \orgaddress{\city{Munich}, \postcode{81675}, \country{Germany}}}

\affil[2]{\orgdiv{Munich Center for Machine Learning (MCML)}, \orgaddress{\city{Munich}, \country{Germany}}}

\affil[3]{\orgdiv{Department of Nuclear Medicine}, \orgname{University Medical Center Hamburg-Eppendorf}, \orgaddress{\city{Hamburg}, \postcode{20246}, \country{Germany}}}

\affil[4]{\orgdiv{Department of Neuroradiology, TUM University Hospital, School of Medicine and Health}, \orgaddress{\city{Munich}, \postcode{81675}, \country{Germany}}}

\affil[5]{\orgdiv{Department of Neurology, TUM University Hospital, School of Medicine and Health}, \orgaddress{\city{Munich}, \postcode{81675}, \country{Germany}}}

\affil[6]{\orgdiv{CompVis}, \orgname{Ludwig-Maximilians-University Munich (LMU)}, \orgaddress{\city{Munich}, \postcode{80539}, \country{Germany}}}

\affil[7]{\orgdiv{Department of Nuclear Medicine, TUM University Hospital, School of Medicine and Health},  \orgaddress{\city{Munich}, \postcode{81675}, \country{Germany}}}


\abstract{
Positron emission tomography (PET) with $^{18}$F-Fluorodeoxyglucose (FDG) is an established tool in the diagnostic workup of patients with suspected dementing disorders. However, compared to the routinely available magnetic resonance imaging (MRI), FDG-PET remains significantly less accessible and substantially more expensive. Here, we present SiM2P, a 3D diffusion bridge-based framework that learns a probabilistic mapping from MRI and auxiliary patient information to simulate FDG-PET images of diagnostic quality. In a blinded clinical reader study, two neuroradiologists and two nuclear medicine physicians rated the original MRI and SiM2P-simulated PET images of patients with Alzheimer's disease, behavioral-variant frontotemporal dementia, and cognitively healthy controls. SiM2P significantly improved the overall diagnostic accuracy of differentiating between three groups from 75.0\% to 84.7\%  ($p<$0.05). Notably, the simulated PET images received higher diagnostic certainty ratings and achieved superior interrater agreement compared to the MRI images. Finally, we developed a practical workflow for local deployment of the SiM2P framework. It requires as few as 20 site-specific cases and only basic demographic information. This approach makes the established diagnostic benefits of FDG-PET imaging more accessible to patients with suspected dementing disorders, potentially improving early detection and differential diagnosis in resource-limited settings. Our code is available at \url{https://github.com/Yiiitong/SiM2P}.
}




\maketitle

\section{Introduction}
\label{intro}

Early and differential diagnosis of dementing disorders remains a significant clinical challenge. At the predementia and mild dementia stages, the sensitivity of routine structural magnetic resonance imaging (MRI) is limited~\cite{dominguez2023review,chouliaras2023use}.
Like regional atrophy observed on MRI, positron emission tomography (PET) using $^{18}$F-Fluorodeoxyglucose ($^{18}$F-FDG) serves as a marker of neurodegeneration by capturing patterns of hypometabolism ~\cite{kato2016brain,strom2022cortical}, however, with a significantly higher sensitivity~\cite{shivamurthy2015brain,minoshima202218f,de2001hippocampal}. This difference is biologically plausible, as neural dysfunction, indexed by hypometabolic areas in FDG-PET, precedes the neuronal death indexed by structural atrophy in MRI~\cite{jack2010hypothetical,jagust2006brain,reiman1998hippocampal}. 
One meta-analysis has reported a sensitivity of 91\% for the diagnosis of Alzheimer's disease (AD) using FDG-PET, compared to 83\% using MRI~\cite{bloudek2011review}.
FDG-PET also demonstrates robust discrimination of AD from other dementia subtypes, with a median sensitivity of 89\%~\cite{fink2020accuracy}, and a specificity exceeding 95\% for differentiating AD from frontotemporal dementia (FTD)~\cite{mosconi2008multicenter,foster2007fdg,panegyres2009fluorodeoxyglucose}. 
Importantly, PET has proven informative even when structural imaging lacks characteristic atrophy~\cite{kerklaan2014added,sala2020brain}. For these reasons, PET is recommended in the diagnostic workup of patients with suspected dementing disorders to support early and differential diagnosis by current guidelines~\cite{frisoni2024european}.

\begin{figure}[hp]
    \centering
    \includegraphics[width=0.9\linewidth]{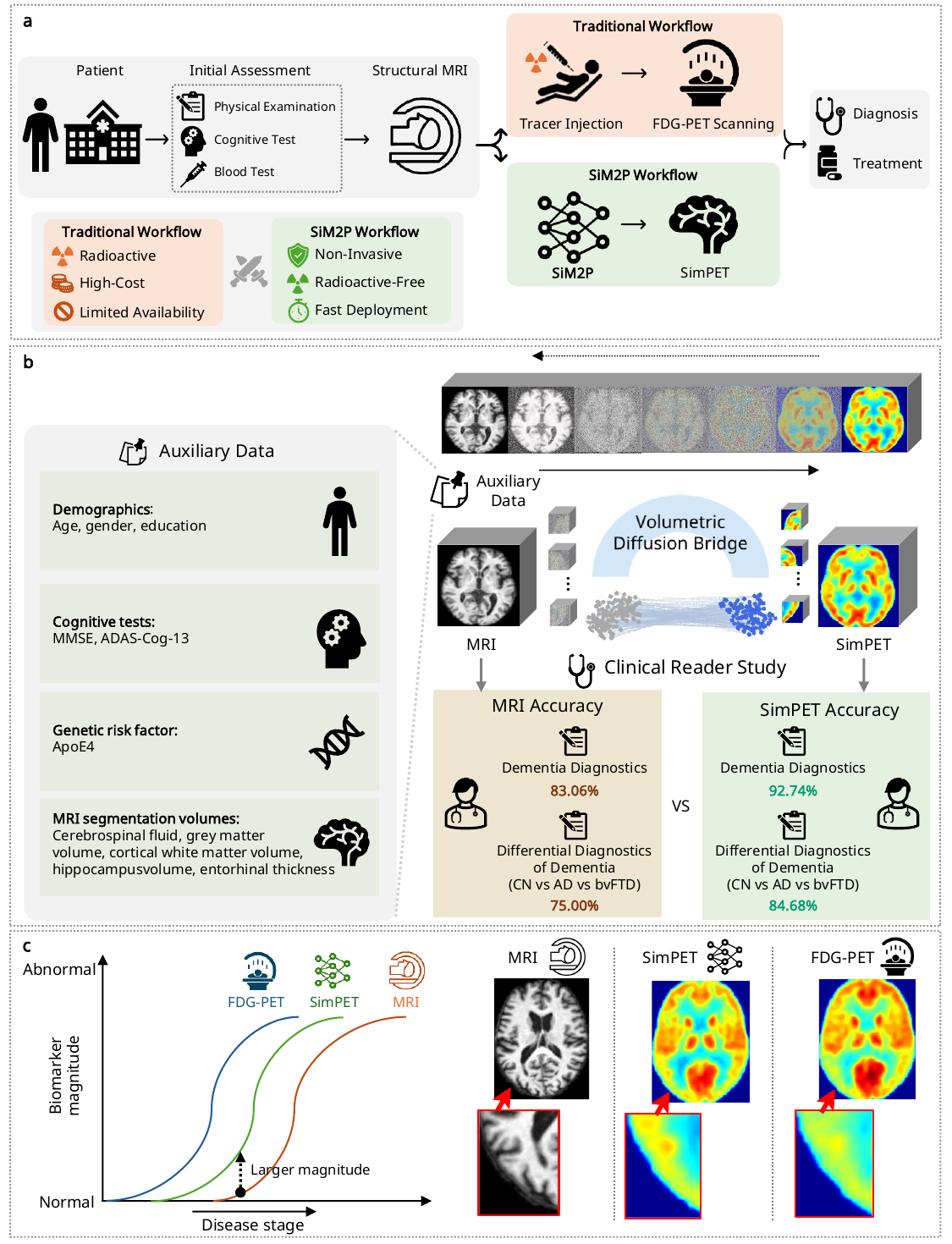}
\caption{\textbf{Overall study design, model pipeline, and evaluation. a}, Clinical context. Clinical diagnostic workup typically involves the structural MRI for assessment of cerebral atrophy, cerebrovascular disease, and exclusion of secondary causes such as tumors. If the diagnosis remains unclear, FDG-PET can be performed in specialized centers. However, the routine use of PET is limited by scanner availability, high costs, and radiation exposure. Our goal is to develop an AI-supported workflow that simulates FDG-PET from routine MRI, enabling PET-informed decision support in settings where PET is unavailable. \textbf{b}, Our model employs a 3D diffusion bridge to simulate PET from structural MRI, conditioned on available auxiliary data such as demographics and MRI-derived segmentation volumes. We validated the diagnostic utility of our simulated PET (SimPET) in a blinded clinical reader study, where SimPET showed a higher accuracy than MRI. \textbf{c}, SiM2P-simulated PET closely resembled disease-specific hypometabolism patterns observed in real FDG-PET and substantially outperformed the biomarker magnitude in MRI.}
\label{fig:overview}
\end{figure}

Despite this value, routine use of PET is constrained by limited scanner availability and high costs~\cite{pet_cost,huang2009whole}. While the associated radiation exposure is less of a concern in patients with advanced age, caution should be exercised in younger subjects~\cite{guedj2022eanm} (Fig.~\ref{fig:overview}a).
MRI, in contrast, is widely available, non-ionizing, and comparatively inexpensive~\cite{pet_ov_mri}. 
To combine the benefits of both, we aim to develop a generative AI model that simulates PET from routine MRI. 
Although both anatomical atrophy in MRI and hypometabolism in PET index the same neurodegeneration process, early brain atrophy in MRI often manifests as subtle changes that may fall below the threshold of human detection~\cite{sadu2025optimizing}. 
Moreover, the mapping from brain anatomy to metabolic function is highly complex and non-linear, varying across brain regions, individuals, and disease stages~\cite{strom2022cortical,fotiadis2024structure}.
Given these challenges, AI models trained with large datasets offer high promise in detecting minute morphological changes beyond human perception, and capturing the intricate, non-linear associations necessary for accurate metabolic inference~\cite{strom2022cortical}. 
A generative formulation is particularly advantageous as it allows label-free training on paired imaging data without reliance on diagnostic supervision, thereby reducing the impact of noisy labels~\cite{rokham2020addressing,azevedo2023identifying}.

Recent progress in diffusion models has transformed conditional image generation, offering state-of-the-art image fidelity and control across diverse domains~\cite{ho2020denoising,dhariwal2021diffusion}. Specifically, the formulation of denoising diffusion bridges provides a principled mechanism for image-to-image translation with improved photorealism and structural faithfulness~\cite{liu2022let,DDBM,shi2023diffusion}. We adapt this advance to 3D medical imaging by introducing \emph{SiM2P}, a novel framework based on a volumetric diffusion bridge (Fig.~\ref{fig:overview}b). 
SiM2P simulates PET scans from structural MRI conditioned on available patient information, coupled with an adaptation workflow to allow for its data-efficient integration into local clinical cohorts for dementia diagnostics.
The resulting simulated PET closely resembles real PET and markedly surpasses the biomarker magnitude on MRI (Fig.~\ref{fig:overview}c).
Unlike prior MRI-to-PET approaches for dementia diagnosis~\cite{gandalf,bpgan,lin2021bidirectional,Li2024pasta,chen2025multi}, which are predominantly limited to AD cases and lack rigorous clinical validation for real-world adoption, SiM2P directly addresses these limitations.
We conducted a blinded clinical reader study to validate its use for dementia subtype diagnosis (Fig.~\ref{fig:overview}b). Furthermore, to facilitate the real-world clinical deployment of SiM2P, we proposed Local-Adapt, an adaptation workflow effectively aligned with site-specific imaging nuances, requiring as low as 20 local cases and minimal patient information.

\begin{table}[h!]
\centering
\caption{Study population. Three independent datasets were used for this study, including ADNI, J-ADNI, and an in-house dataset from our hospital. The $p$-value for each dataset indicates the statistical significance of intergroup differences per column. We used one-way analysis of variance (ANOVA) and two-sided $\chi^2$ tests for continuous and categorical variables, respectively. (CN: healthy controls, AD: Alzheimer's disease, MCI: mild cognitive impairment, FTLD: frontotemporal lobar degeneration, ND: subjects without evidence for a neurodegenerative disease, s.d.: standard deviation.)}
\label{tab:demographics}
\renewcommand{\arraystretch}{1.5}
{\fontsize{6.5pt}{7.5pt}\selectfont
    \begin{tabular}{lcccccc}
        \toprule
        Dataset  & Age (y),  & Male,  & Edu (y), & MMSE, & ADAS13,  & APOE4  \\
        (group) & mean $\pm$ s.d. & n (\%) & mean $\pm$ s.d. & mean $\pm$ s.d. & mean $\pm$ s.d. & (0/1/2), n \\
        \midrule
        \textbf{ADNI} \\
         CN [n=379] & 73.47 $\pm$ 5.94 & 186 (49.1\%) & 16.38 $\pm$ 2.73 & 29.00 $\pm$ 1.19 & 9.27 $\pm$ 4.33 & 273/95/9 \\
         MCI [n=611] & 72.31 $\pm$ 7.31 & 358 (58.6\%) & 16.09 $\pm$ 2.73 & 27.82 $\pm$ 1.74 & 15.61 $\pm$ 6.69 & 308/237/66 \\
         AD [n=257] & 74.42 $\pm$ 7.91 & 153 (59.5\%) & 15.41 $\pm$ 2.84 & 23.16 $\pm$ 2.18 & 30.78 $\pm$ 8.13 & 79/121/52 \\
         $p$-value & 1.67$\times 10^{-4}$ & 5.71$\times10^{-3}$ & 6.62$\times 10^{-5}$ & 6.94$\times10^{-256}$ & 1.70$\times10^{-25}$ & 1.62$\times10^{-235}$ \\
        \midrule
        \textbf{J-ADNI} \\
         CN [n=104] & 67.88 $\pm$ 5.40 & 53 (51.0\%) & 13.75 $\pm$ 2.81 & 29.15 $\pm$ 1.22 & 7.67 $\pm$ 4.25 & 80/22/2 \\
        MCI [n=131] & 72.31 $\pm$ 5.76 & 66 (50.4\%) & 13.38 $\pm$ 2.91 & 26.31 $\pm$ 1.64 & 19.30 $\pm$ 6.63 & 57/66/7 \\
        AD [n=84] & 74.01 $\pm$ 6.52 & 39 (46.4\%) & 12.35 $\pm$ 2.85 & 22.54 $\pm$ 1.77 & 27.47 $\pm$ 5.54 & 33/35/16 \\
        $p$-value & 3.75$\times 10^{-12}$ & 8.00$\times 10^{-1}$ & 3.00$\times 10^{-3}$ & 6.00$\times10^{-90}$ & 7.32$\times10^{-10}$ & 5.72$\times10^{-72}$ \\
        \midrule
        \textbf{In-house} \\
        ND [n=143] & 64.15 $\pm$ 9.97 & 76 (53.1\%) & -- & -- & -- & -- \\
        AD [n=110] & 67.25 $\pm$ 8.37 & 55 (50.0\%) & -- & -- & -- & -- \\
        FTLD [n=70] & 65.37 $\pm$ 9.16 & 41 (58.6\%) & -- & -- & -- & -- \\
        $p$-value & 3.21$\times 10^{-2}$ & 5.32$\times 10^{-1}$ & -- & -- & -- & -- \\
        \bottomrule
    \end{tabular}}
\end{table}

\section{Results}

\subsection{Characteristics of the study cohorts}

We leveraged the SiM2P framework with multimodal data from diverse cohorts (Table~\ref{tab:demographics}, Extended Data Table~\ref{tab:mri_measures}) to simulate accurate PET scans from routine MRIs and auxiliary patient information.
Model development and pre-training leveraged the data from the ADNI~\cite{adni} and J-ADNI~\cite{jadni} cohorts. The independent in-house data were then used for model adaptation, with its held-out test set preserved for final evaluation in the clinical reader study.
Furthermore, all three cohorts were merged to provide a comprehensive evaluation of image quality against state-of-the-art methods.

\subsection{Clinical reader study}
\label{sec:clinical_reader_study}

\begin{figure}[hp]
    \centering
    \includegraphics[width=0.98\linewidth]{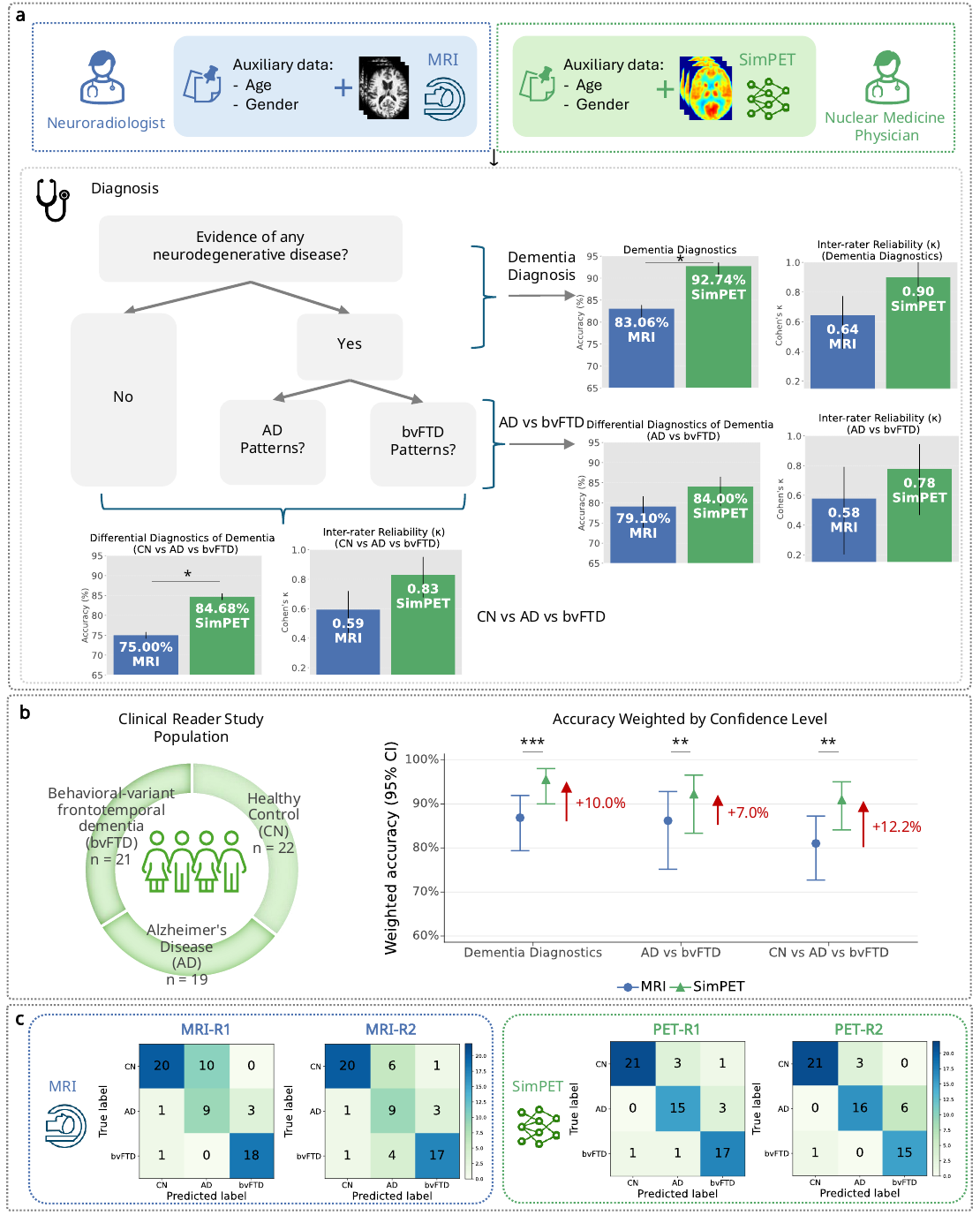}
    \caption{\textbf{Clinical reader study pipeline and results. a}, The study involves a two-stage diagnostic workflow for both neuroradiologists (for MRI) and nuclear medicine physicians (for simulated PET, SimPET in short). We report the diagnostic accuracy alongside interrater reliability (using Cohen's $\kappa$) for three tasks: dementia disorders diagnosis, differential diagnosis of dementia disorders for either AD-versus-bvFTD or CN-versus-AD-versus-bvFTD, with error bars representing within-rater standard deviation and 95\% CI, respectively. Significance levels using McNemar's test are denoted on top as *$P<$ 0.05. \textbf{b}, Study population and diagnostic accuracy weighted by each rater’s confidence level for MRI and SimPET across three tasks. Error bars indicate the 95\% CI. The percent increase in mean performance gained by SimPET is indicated alongside each task. Significance levels using the Wilcoxon signed-rank test are denoted on top as *$P<$ 0.05, **$P<$  0.01, ***$P<$ 0.001. \textbf{c}, Confusion matrices of all raters for MRI (left) and SimPET (right) across three labels.}
    \label{fig:clinical_study}
\end{figure}

\subsubsection{Diagnostic performance of simulated PET compared to MRI}

We reported the diagnostic performance of simulated PET and MRI in both dementia diagnosis, i.e., identifying the presence of any dementia disorder, and differential diagnosis, i.e., distinguishing between AD and behavioral-variant FTD (bvFTD) (Fig.~\ref{fig:clinical_study}a), as accuracy with 95\% Wilson confidence intervals and per-class sensitivity and specificity (Extended Data Fig.~\ref{fig:sen_spec}). In the former task, simulated PET achieved an average accuracy of 92.74\% (95\% CI: 86.78\% to 96.13\%), with 91.94\% from PET-R1 and 93.55\% from PET-R2. In contrast, MRI yielded an average accuracy of 83.06\% (95\% CI: 75.49\% to 88.65\%), with 80.65\% from MRI-R1 and 85.48\% from MRI-R2. Simulated PET led to a mean absolute improvement of 9.69 percentage points (an 11.7\% relative gain) over MRI ($p<0.05$).

In the AD‐versus‐bvFTD differentiation, simulated PET achieved 84.68\% (95\% CI: 77.31\% to 89.97\%), with 86.49\% for PET-R1 and 81.58\% for PET-R2, which also surpassed the average MRI accuracy of 79.10\% (95\% CI: 67.93\% to 87.12\%), with 84.38\% for MRI-R1 and 74.29\% for MRI-R2, corresponding to a mean absolute improvement of 4.70 points (a 5.9\% relative gain) ($p\geq0.05$).
For the three‐class differential diagnosis between CN, AD, and bvFTD, simulated PET reached an average of 84.68\% (95\% CI: 77.31\% to 89.97\%), with 85.48\% for PET-R1 and 83.87\% for PET-R2, yielding a mean absolute gain of 9.68 points (a 12.9\% relative gain) over MRI's average of 75.00\% (95\% CI: 66.71\% to 81.79\%), with 75.81\% for MRI-R1 and 74.19\% for MRI-R2 ($p<0.05$).

\subsubsection{Confidence-weighted diagnostic accuracy}

To better reflect diagnostic uncertainty in the rating results, we prospectively recorded each rater's per-case diagnostic confidence and calculated a confidence-weighted accuracy (Section~\ref{sec:study_design}), providing a more informative measure of diagnostic performance.
As shown in Fig.~\ref{fig:clinical_study}b, the higher diagnostic accuracy gained from simulated PET was consistent and even amplified under this metric. For dementia diagnosis, raters using simulated PET achieved a confidence-weighted accuracy of 95.53\% (95\% CI: 90.00\% to 98.06\%), a significant gain ($p<0.001$) over MRI's 86.88\% (95\% CI: 79.38\% to 91.93\%), representing a mean absolute gain of 8.65 percentage points (a 10.0\% relative gain). 
For the binary AD‐versus‐bvFTD differentiation, simulated PET's weighted accuracy was 92.22\% (95\% CI: 83.33\% to 96.57\%), compared to MRI's 86.21\% (95\% CI: 75.17\% to 92.81\%), gaining a 6.0\% absolute accuracy increase (a 7.0\% relative gain) ($p<0.005$).
Finally, in the three‐class differential diagnosis, simulated PET reached 90.94\% (95\% CI: 84.10\% to 95.01\%), resulting in a 9.9\% absolute accuracy gain (a 12.2\% relative gain) significantly ($p<0.005$) higher than MRI's 81.04\% (95\% CI: 72.71\% to 87.27\%). 
We demonstrated further analysis on the rater's diagnostic correctness with regard to the confidence level in Extended Data Fig.~\ref{fig:confidence_analysis}.
Simulated PET significantly outperformed MRI ($p<0.005$) across all three tasks, demonstrating simultaneous improvements in overall diagnostic accuracy and the reliability of high-confidence decisions.

\subsubsection{Interrater reliability}

Consistently, simulated PET yielded higher interrater agreement than MRI across all diagnostic tasks (Fig.~\ref{fig:clinical_study}a).
For the detection of dementia disorders, simulated PET achieved excellent agreement (Cohen’s kappa statistic $\kappa$ = 0.899, 95\% CI: 0.769 to 1.0), compared to the moderate agreement observed with MRI ($\kappa$ = 0.644, 95\% CI: 0.419 to 0.808).
Similarly, simulated PET demonstrated substantially higher agreement for differentiating AD from bvFTD ($\kappa$ = 0.778, 95\% CI: 0.494 to 0.912), and for the three-class differential diagnosis ($\kappa$ = 0.829, 95\% CI: 0.704 to 0.950). In both tasks, MRI showed only moderate agreement ($\kappa$ = 0.578, 95\% CI: 0.160 to 0.814, and $\kappa$ = 0.595, 95\% CI: 0.395 to 0.746, respectively).

\begin{figure}[hp]
    \centering
    \includegraphics[width=0.99\linewidth]{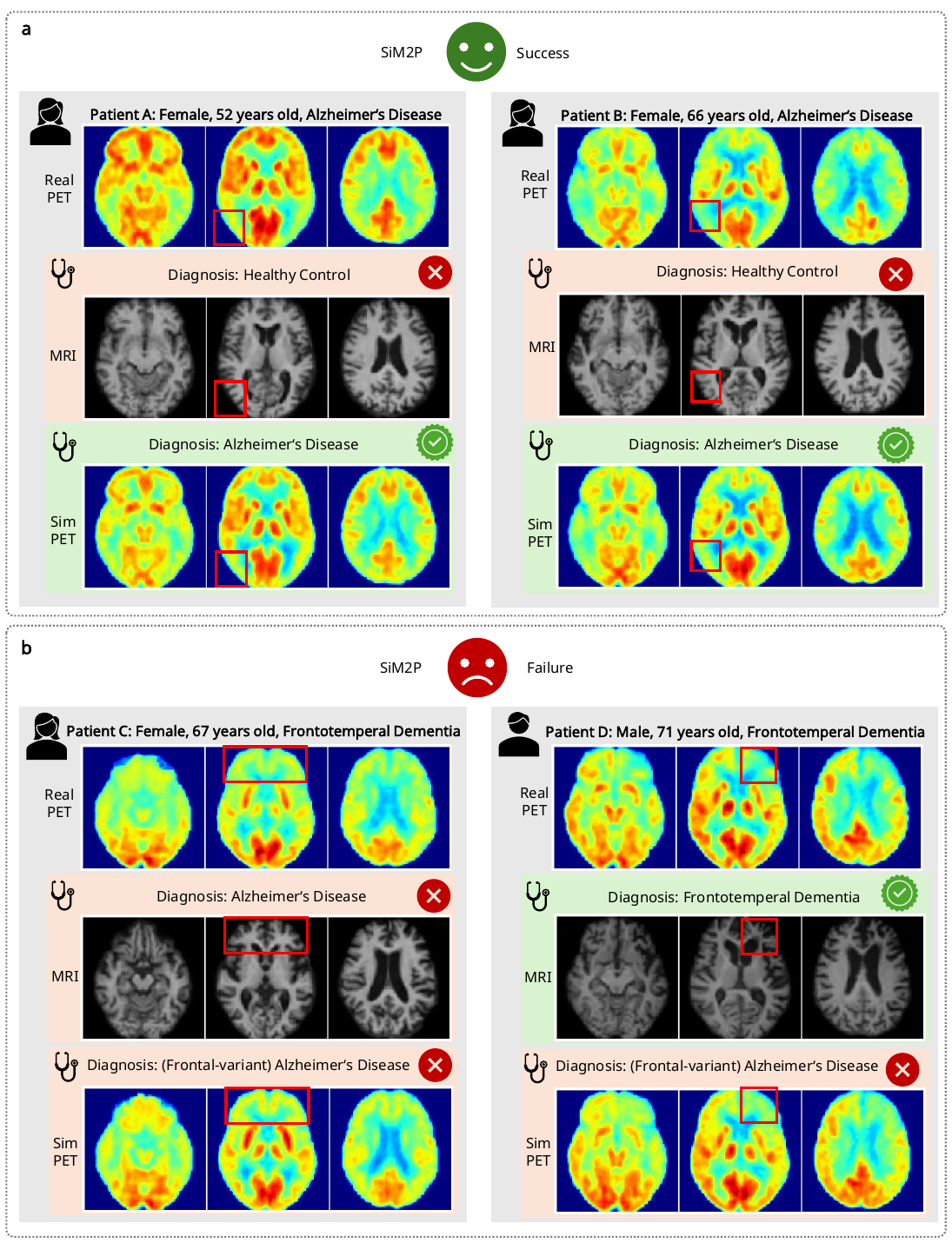}
    \caption{\textbf{Representative success and failure cases of SiM2P in the clinical reader study. a}, Two success cases in which subtle MRI atrophy patterns led to incorrect diagnosis as healthy control subjects, whereas simulated PET (SimPET) reproduced the temporoparietal hypometabolism seen on the real PET, enabling correct diagnosis of Alzheimer’s disease. \textbf{b}, Two failure cases in which overlapping metabolic patterns of frontal-variant AD and behavioral-variant FTD resulted in misdiagnosis of AD, despite SimPET closely matching the frontal-lobe hypometabolism observed on real PET. Abnormal regions are highlighted with red boxes on the middle slice.}
    \label{fig:example_cases}
\end{figure}

\subsubsection{Analysis of success and failure diagnostic cases}

Simulated PET provided higher sensitivity for dementia disorders, particularly AD. 
Fig.~\ref{fig:example_cases}a shows two representative AD cases. Here, MRI displayed only subtle and less discernible atrophy, leading to misdiagnosis of CN. 
Real PET and simulated PET both revealed clear hypometabolism in the temporoparietal regions, enabling the correct diagnosis of AD.
Overall, seven subjects who were misclassified as CN by neuroradiologists based on MRIs were correctly identified as AD patients by both nuclear medicine physicians using simulated PET.

Misdiagnoses with simulated PET frequently occurred in cases with overlapping disease types. For example, three bvFTD patients were misdiagnosed as AD by both nuclear medicine physicians using simulated PET. 
Closer inspection revealed that these cases were predominantly frontal-variant AD (fvAD), a subtype difficult to distinguish from bvFTD due to their substantial overlap in metabolic patterns, particularly in the frontal cortex~\cite{wong2016comparison,brown2023frontal}.
As illustrated by two such cases in Fig.~\ref{fig:example_cases}b, simulated PET faithfully reproduced the frontal hypometabolism seen on the real PET. While this led both nuclear medicine physicians to express high confidence in the detection of dementia disorders, they marked the pattern as low confidence for an AD subtype, leading to a tentative AD diagnosis. One rater additionally commented that these cases resembled fvAD.
In total, such misdiagnosis occurred in three out of twenty-one (14.3\%) total bvFTD cases for PET-R1, and six out of twenty-one (28.6\%) cases for PET-R2. 
These cases indicate that simulated PET could reveal disease-related abnormalities that are less obvious in MRI, while failure cases largely arise from closely resembling disease types.

\subsection{Comparisons with other generative methods}
\label{sec:comparison_results}

The resulting simulated PET scans from SiM2P closely resemble real PET, preserving both anatomical fidelity and disease-characteristic hypometabolism across different disease groups (Extended Data Fig.~\ref{fig:qualitative_vis}).
We further benchmarked SiM2P against four state-of-the-art generative models, namely Pix2Pix~\cite{pix2pix}, ResViT~\cite{resvit}, BBDM~\cite{bbdm}, and PASTA~\cite{Li2024pasta}, both qualitatively and quantitatively.

\begin{figure}[hp]
    \centering
    \includegraphics[width=0.86\linewidth]{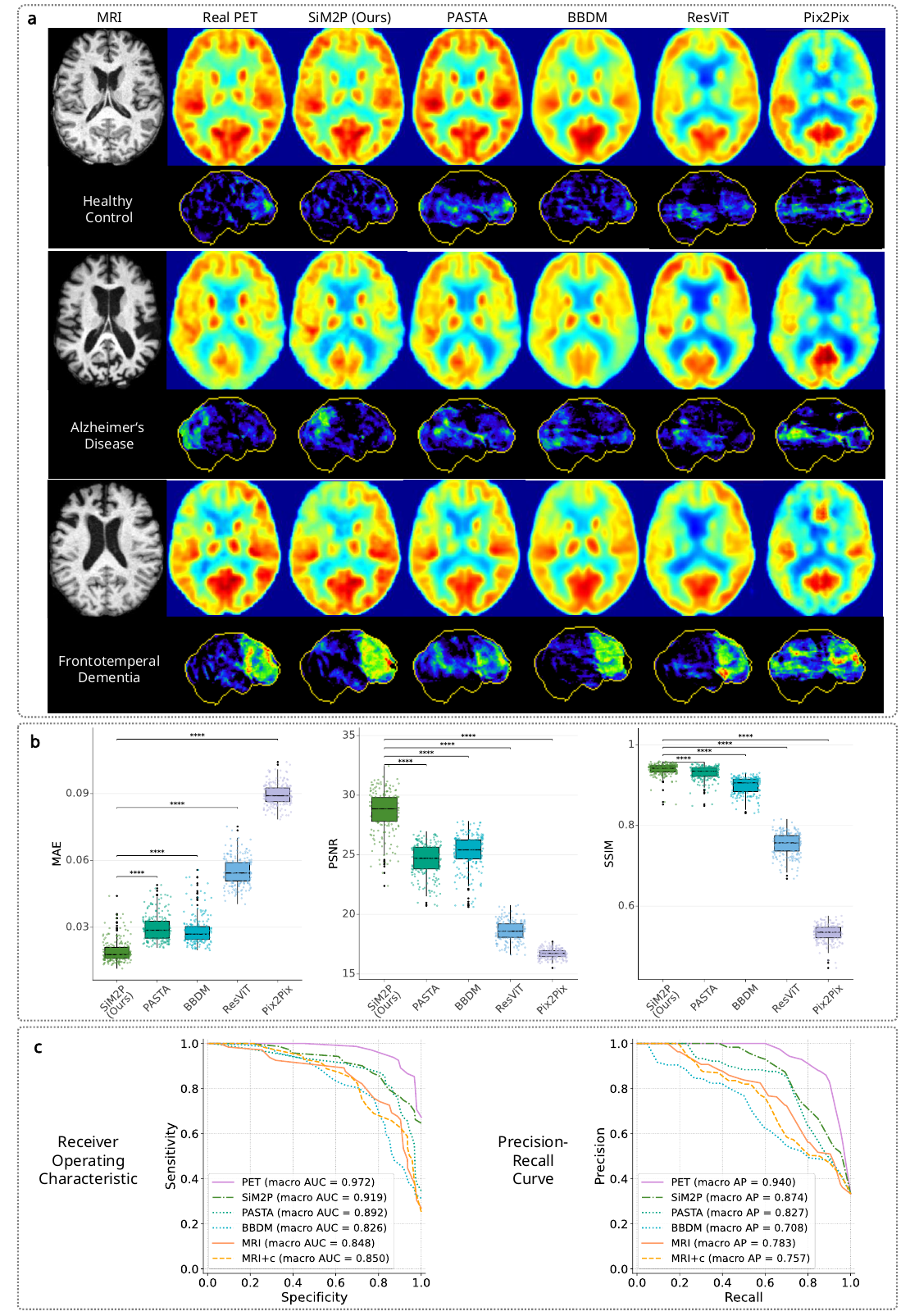}
    \caption{\textbf{Qualitative and quantitative comparison. a}, Visual comparison of PET and corresponding 3D-SSP maps (global-normalized negative z-score maps from right hemisphere lateral view) generated by SiM2P and competing methods, alongside real PET and MRI across CN, AD, and bvFTD cases. \textbf{b}, Quantitative evaluation using MAE($\downarrow$), PSNR($\uparrow$), and SSIM($\uparrow$). Each boxplot shows the median line and interquartile range (IQR) box, with whiskers extending to the most extreme values within 1.5 $\times$ IQR; individual data points are overlaid. All pairwise comparisons were statistically significant with ****$P<$ 0.0001, using the one-tailed Wilcoxon signed-rank test without corrections made for multiple comparisons. \textbf{c}, Macro-averaged ROC and PR curves for a downstream automated classification task. The performance of simulated PET from SiM2P is benchmarked against MRI, MRI with auxiliary clinical data (MRI+c), real PET, and generated PET from other diffusion model-based methods.}
    \label{fig:comparison_results}
\end{figure}

\subsubsection{Qualitative evaluation}

For a qualitative assessment, we compared representative PET generation results from SiM2P with those from competing methods, alongside corresponding 3D-SSP visualizations~\cite{minoshima1995diagnostic} (Sec.~\ref{sec:method_3dssp}) for CN, AD, and bvFTD cases (Fig.~\ref{fig:comparison_results}a). 
Visual inspection demonstrated that SiM2P consistently produced PET with higher fidelity and a closer match to the real PET than all other methods.  
For example, in AD cases, SiM2P accurately reproduced the characteristic hypometabolism in the temporoparietal lobes, a hallmark region strongly associated with AD~\cite{nestor2003limbic}, closely mirroring the real PET. Similarly, for bvFTD, SiM2P simulated PET scans with marked frontal lobe hypometabolism and prominent left-right asymmetric metabolic patterns in the frontal lobes, closely matching the real PET.
In contrast, other diffusion model-based methods, PASTA and BBDM, tended to produce overly smooth outputs that lacked fine-grained details. While these models preserved structural information, they only partially captured the pathological patterns seen in real PET. 
The rest two methods showed even greater limitations. ResViT captured certain pathological features but compromised anatomical accuracy, while Pix2Pix exhibited substantial artifacts and structural inconsistencies.

\subsubsection{Quantitative evaluation}

We quantitatively evaluated the quality and fidelity of the simulated PET using several key metrics: mean absolute error (MAE$\downarrow$), mean squared error (MSE$\downarrow$), peak signal-to-noise ratio (PSNR$\uparrow$), and structure similarity index (SSIM$\uparrow$). 
As shown in Fig.~\ref{fig:comparison_results}b and Extended Data Table~\ref{tab:quantitative_results_comparison}, SiM2P significantly outperformed all baseline methods across every metric ($p<0.0001$) on the merged dataset with ADNI~\cite{adni}, J-ADNI~\cite{jadni}, and in-house data. Our method achieved the lowest MAE (0.0192 $\pm$ 0.0049) and MSE (0.0015 $\pm$ 0.0007), while reaching the highest PSNR (28.59 $\pm$ 1.73) and SSIM (0.9393 $\pm$ 0.0139). The diffusion model-based methods, PASTA and BBDM, consistently demonstrated the second-best performance.  In contrast, the other GAN-based baselines, particularly Pix2Pix, could not reach on-par performance.

\subsubsection{Automated classification results with simulated PET}
\label{sec:automated_classification}

To benchmark the diagnostic utility of the simulated PET, we performed a downstream automated classification task using different modalities as input. 
These modalities included MRI, MRI with auxiliary clinical data (MRI+c), original PET, and simulated PET from SiM2P and competing diffusion model-based methods, PASTA and BBDM. 
This deep learning-based downstream task was for the three-class differential diagnosis of dementia on the in-house dataset (Sec.~\ref{sec:automated_classification}).
We presented the macro-averaged Receiver Operating Characteristic (ROC) and Precision-Recall (PR) curves along with their corresponding Area Under Curve (AUC) and Average Precision (AP) scores (Fig.~\ref{fig:comparison_results}c).
Detailed per-class ROC and PR curves and their micro-averaged results are available in Extended Data Fig.~\ref{fig:macro_micro_AP_AUC_plots}. 

As shown by the ROC plot, the simulated PET from SiM2P closely approximated the performance of real PET. Our method achieved a macro-averaged AUC of 0.919 (95\% CI: 0.848 to 0.968), approaching the 0.972 (95\% CI: 0.934 to 0.994) obtained by the original PET data. Moreover, SiM2P numerically outperformed other methods, such as PASTA (macro AUC = 0.892, 95\% CI: 0.776 to 0.955) and BBDM (macro AUC = 0.826, 95\% CI: 0.743 to 0.894), as well as MRI-based inputs (MRI: macro AUC = 0.848, 95\% CI: 0.722 to 0.932; MRI+c: macro AUC = 0.850, 95\% CI: 0.760 to 0.918). 

Similar trends were observed in the PR curves, where SiM2P demonstrated a macro-averaged AP of 0.874 (95\% CI: 0.764 to 0.936), numerically higher than other methods (PASTA: macro AP = 0.827, 95\% CI: 0.702 to 0.910; BBDM: 0.708, 95\% CI: 0.587 to 0.811) and MRI-based inputs (MRI: 0.783, 95\% CI: 0.660 to 0.880; MRI+c: 0.757, 95\% CI: 0.625 to 0.850), approaching the level of original PET imaging (macro AP = 0.940, 95\% CI: 0.857 to 0.966). 
Overall, the classification performance using SiM2P significantly outperformed MRI, MRI+c, and BBDM ($p< 0.05$).

\subsection{Clinical deployment on the local cohort}
\label{sec:local_adapt}

\begin{figure}[hp]
    \centering
    \includegraphics[width=0.94\linewidth]{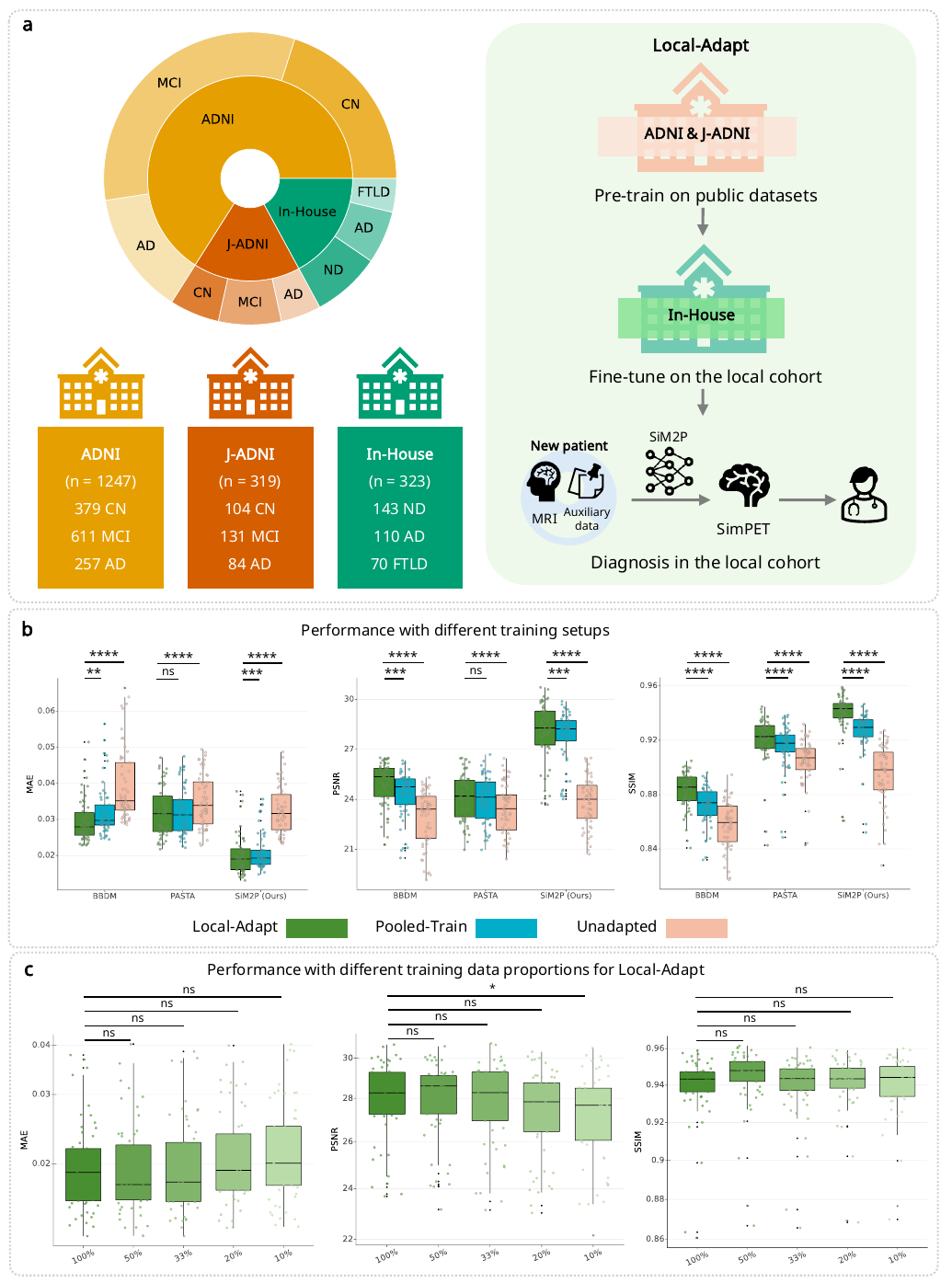}
    \caption{\textbf{Local-Adapt workflow and results. a}, Schematic of the two-stage Local-Adapt pipeline, with large publicly available datasets for pre-training and the smaller local data for fine-tuning. \textbf{b}, Performance comparison across three training setups: Local-Adapt, Pooled-Train, and Unadapted, for SiM2P, PASTA, and BBDM models. \textbf{c}, SiM2P's performance with varying proportions of fine-tuning data in Local-Adapt. Both \textbf{b} and \textbf{c} measure the performance using MAE($\downarrow$), PSNR($\uparrow$), and SSIM($\uparrow$). Boxplots show the median line and interquartile range (IQR) box, with whiskers extending to the most extreme values within 1.5 $\times$ IQR; individual data points are overlaid. Pairwise statistical comparisons were conducted using a one-tailed Wilcoxon signed-rank test without corrections for multiple comparisons. Significance levels are denoted as ns (not significant) for $P\geq$ 0.05, *$P<$ 0.05, **$P<$ 0.01, ***$P<$ 0.001, ****$P<$ 0.0001.}
    \label{fig:local_adapt}
\end{figure}

For the clinical deployment of SiM2P, we developed a lightweight adaptation workflow, Local-Adapt, that enabled SiM2P to be efficiently deployed at local clinical cohorts while preserving high generation quality. 
The SiM2P model was first pre-trained on publicly available datasets, ADNI~\cite{adni} and J-ADNI~\cite{jadni}, to learn generalizable structure-to-function relationships. It was then rapidly fine-tuned on a small local clinical dataset to capture site-specific scanner and protocol characteristics (Fig.~\ref{fig:local_adapt}a). While pre-training took a considerable amount of time, normally three to four days, the fine-tuning phase only took a few hours to complete. Importantly, the fine-tuning stage did not require the full set of auxiliary data used during pretraining; successful deployment on our local cohort was achieved using only age, gender, and MRI-derived segmentation volumes that are naturally available from the MRI.

We benchmarked Local-Adapt against two common training strategies (Fig.~\ref{fig:local_adapt}b): Pooled-Train (full training on the collection of public and local data) and Unadapted (direct use of a pre-trained model on the local data). Across three diffusion model-based methods, including SiM2P, PASTA~\cite{Li2024pasta}, and BBDM~\cite{bbdm}, Local-Adapt delivered the best overall performance. It significantly outperformed Unadapted ($p<0.0001$), and Pooled-Train in SiM2P and BBDM ($p<0.01$). Although Pooled-Train could achieve comparable MAE scores across methods, Local-Adapt offered a substantially faster and more flexible alternative to the computationally intensive Pooled-Train strategy.

Notably, SiM2P’s generation quality remains robust even under severely limited data resources in the Local-Adapt fine-tuning stage. We systematically reduced the fine-tuning data from the full in-house training set (196 cases) down to 50\% (98 cases), 33\% (65 cases), 20\% (39 cases), and even 10\% of the data (19 cases). 
As shown in Fig.~\ref{fig:local_adapt}c, performance metrics like MAE and SSIM showed no significant differences across these settings ($P\geq0.05$). This resilience highlights the practicality of SiM2P for local clinical deployment, where access to a large amount of patient data is often restricted.

\subsection{Impact of auxiliary data}
\label{sec:impact_clinical_data}

To quantify the contribution of auxiliary patient information in SiM2P, we experimented with excluding all auxiliary data and trained the model using only the MRI scans as input.
This exclusion led to a significant performance drop on the validation set. The MAE increased from $0.0196 \pm 0.0054$ to $0.0212 \pm 0.0061$ ($p<0.005$), while the PSNR decreased from $28.62 \pm 1.83$ to $27.72 \pm 1.87$ ($p<0.0001$).

We further analyzed the performance difference using each auxiliary variable as input compared to the baseline configuration, in which all variables were retained (Sec.~\ref{sec:method_auxiliary_influence}). As detailed in Extended Data Fig.~\ref{fig:clinical_data_influence}, cognitive scores (ADAS-Cog-13, MMSE), genetic risk factor ApoE4, and education level consistently outperformed other demographics and MRI segmentation volumes, yielding a smaller performance difference in MAE and SSIM. Among the MRI-derived segmentation volumes, white matter volume was found to be more informative.
These results highlight the added value of integrating auxiliary clinical and anatomical features, which provide complementary information and enable more accurate PET simulation.


\section{Discussion}

We introduced SiM2P, a 3D diffusion bridge-based framework that conditions on routine structural MRI and auxiliary patient information to simulate clinical-grade FDG-PET images with rigorous evaluation.
Hypometabolic areas in the simulated PET images closely resembled the known disease-specific patterns of neurodegeneration. In a blinded clinical reader study, simulated PET images showed superior diagnostic accuracy compared to structural MRI. Further, simulated PET images showed high sensitivity to pathology and outperformed state-of-the-art competing methods of automatic diagnosis. This framework offers access to the established diagnostic benefits of PET imaging, thereby accelerating early detection and differential diagnosis of dementing disorders.

In the clinical reader study, simulated PET from SiM2P achieved significantly higher diagnostic accuracy over MRI in both the initial diagnosis of dementia disorders and the critical differential diagnosis between CN, AD, and bvFTD. When incorporating per-case diagnostic certainty, the performance gains of simulated PET images were further amplified and significant. This confidence-weighted analysis provided a more clinically realistic measure of performance, as decisions made with higher certainty were given more weight.
Under this metric, simulated PET continued to outperform MRI, leading to a 10.0\% higher accuracy in dementia diagnosis, and a 12.2\% gain in the differential diagnosis of dementia. This suggests that SiM2P-simulated PET not only leads to more accurate, but also to more confident diagnoses.
In addition, simulated PET achieved substantially higher interrater reliability compared to MRI.
This improved interrater agreement suggests that simulated PET provides clearer, more distinct disease signatures for greater diagnosis consensus.
Such a reduction in diagnostic ambiguity fosters higher reliability in clinical decision-making.

Our case-level analysis on diagnostic success and failure cases provided deeper insights into the strengths and limitations of the simulated PET.
SiM2P proved most valuable in cases with subtle or non-specific pathological atrophy on MRI, where it revealed characteristic hypometabolism that led to the correction of MRI-based misdiagnoses. This highlights the higher sensitivity that metabolic information provides in early or atypical presentations of dementia disorders. 
Failure cases often happened in cases with significant pathological overlap, particularly between frontal-variant AD and bvFTD~\cite{wong2016comparison,brown2023frontal}.
In such cases, misdiagnosed simulated PET still faithfully reproduced the frontal hypometabolic patterns seen on the original PET, indicating an inherent challenge in distinguishing these conditions. This suggests that the misdiagnoses might not be only due to imperfect performance of the simulation process but also due to an inherent spatial overlap of disease patterns.

Methodologically, SiM2P is powered by a conditional diffusion bridge with a 3D diffusion Transformer as backbone. 
Recent advances in generative modeling~\cite{liu2022let,DDBM,songscore2021} show that the design space of denoising diffusion models can be made highly flexible using diffusion bridge processes. These processes can be seen as stochastic trajectories with fixed endpoints~\cite{sarkka2019applied,heng2021simulating,shi2023diffusion}.
Within this framework, SiM2P naturally frames the task of MRI to PET simulation as constructing an appropriate diffusion bridge process (i.e., a probabilistic mapping) from a subject's structural anatomy to the metabolic signals. This direct mapping from the full 3D MRI volume anchors the generation trajectory to subject-specific anatomy, preserving fine-grained details with enhanced pathological alignment.
By leveraging a 3D diffusion Transformer as its backbone, SiM2P can scale effectively with increasing resources. This enhanced capacity is crucial for learning the complex, fine-grained relationships required for a robust structural-to-functional mapping from MRI to PET. 
It also allows the model to be readily conditioned on auxiliary patient information, such as demographics, cognitive scores, and genetic risk factors. These low-dimensional priors provide complementary information to improve the structural-to-functional mapping and personalize the simulated PET with patient-specific information.

Compared to other state-of-the-art generative models~\cite{pix2pix,resvit,bbdm,Li2024pasta}, SiM2P produced PET scans with superior fine-grained and accurate metabolic details, significantly outperforming all competing methods in various image quality metrics.
Qualitative assessments alongside 3D-SSP visualizations~\cite{minoshima1995diagnostic} confirmed that the disease-specific hypometabolic patterns on simulated PET from SiM2P closely matched those on real PET.
The diagnostic utility of SiM2P-simulated PET was further validated using an automated classifier for the differential diagnosis of dementia. 
SiM2P yielded significantly higher accuracy than MRI and surpassed the performance of the generated PET from competing methods. This demonstrates that SiM2P not only achieves high image quality but also preserves critical disease-relevant metabolic patterns.
Together, these results indicate that SiM2P is not merely producing realistic-looking PET scans. It is learning to induce accurate metabolic signals by constructing a reliable and robust structural-to-functional mapping.

For local clinical deployment, pre-trained SiM2P models can be rapidly tailored to site-specific data through a practical adaptation workflow, Local-Adapt. The scarcity of paired MRI-PET data at individual sites is commonly a major translational barrier to the clinical deployment of AI-supported models. To address this challenge, SiM2P incorporated Local-Adapt, a highly data-efficient workflow that maintained high-quality and accurate PET simulation even when fine-tuned with as few as 20 local samples. Importantly, Local-Adapt did not require the full set of auxiliary information used during pre-training; instead, age, gender, and routine MRI-derived segmentation volumes were sufficient to recover clinically meaningful PET simulation, confirmed by the expert reading on our local cohort. 
This suggests a highly feasible path for SiM2P's local adoption. Clinical sites can calibrate publicly pre-trained SiM2P models with a minimal number of paired cases to account for scanner- and population-specific characteristics. This makes it possible to integrate reliable simulated PET for dementia diagnosis in individual clinical cohorts where traditional PET scanning is unavailable or impractical, effectively reducing costs and radiation exposure.

While SiM2P shows significant promise for providing low-cost, radiation-free diagnostic assistance with simulated PET, some limitations remain. 
Since our study demonstrated its diagnostic utility within our cohort, a crucial next step is to conduct prospective, multi-institutional studies to confirm its generalizability. As currently limited by data privacy regulations, we anticipate this can be partially mitigated by making our pre-trained model publicly available, enabling rapid, site-specific adaptation to various local cohorts.
From a technical perspective, the computational demands of training a 3D diffusion bridge model are substantial, requiring significant time and resources (around 96 GPU hours).
This challenge is partially addressed by our Local-Adapt workflow to enable local cohorts with limited resources to rapidly adapt our pre-trained models. Future research is needed to optimize these processes and make the framework even more accessible for clinical practice. Also, while simulated PET can assist physicians in diagnostic decision-making, its capacity for the absolute quantification of biomarkers requires further validation. Finally, defining the appropriate interpreting specialty of simulated PET images presents a trade-off. While nuclear medicine physicians offer the definitive expertise, as used herein, neuroradiologists are logistically favored as they are responsible for the acquisition of MRI data to be used for simulation. Leveraging nuclear medicine physicians for the task would necessitate a telemedicine infrastructure. 

Looking ahead, SiM2P provides a generalizable framework for extending simulated FDG-PET scans to other neurodegenerative disorders, e.g., movement disorders, or different PET tracers, e.g., amyloid, tau. Multimodal conditioning with cerebrospinal fluid, e.g., p-tau, t-tau, may enhance subtype specificity and prognostic values.

In conclusion, SiM2P addresses a critical gap in dementia diagnostics by bridging readily available structural MRI with more sensitive metabolic information from PET. SiM2P generates simulated PET images that are not only diagnostically accurate but also adaptable in local clinical centers. 
Our results indicate that SiM2P is not only capable of capturing subtle anatomical changes that often escape human detection in early disease stages, but also of learning the complex, non-linear structural-to-functional relationships critical for metabolic inference.
Importantly, this generative formulation allows it to learn this intricate signal mapping directly from paired imaging data, mitigating the impact of potentially noisy or imperfect diagnostic labels in medical imaging datasets~\cite{rokham2020addressing,azevedo2023identifying}.
This represents a significant step toward widening PET-informed decision support, making the established PET diagnostic insights more accessible for routine dementia care.

\section{Methods}
\label{method}

\subsection{Datasets}

We created our merged dataset with 1,860 paired T1-weighted magnetic resonance imaging (MRI) scans and $^{18}$F-fluorodeoxyglucose-positron emission tomography ($^{18}$F-FDG PET) subjects from three cohorts. The cohorts include the Alzheimer’s disease neuroimaging initiative (ADNI)~\cite{adni} dataset ($n = 1,247$), the Japanese Alzheimer’s Disease Neuroimaging Initiative
(J-ADNI)~\cite{jadni} database ($n = 319$), and our in-house clinical dataset ($n = 323$) from TUM University Hospital (TUM Klinikum), Munich, Germany. This merged dataset incorporates diverse patient demographics and geographical origins spanning three distinct continents.

The ADNI was launched in 2003 as a public-private partnership, led by Principal Investigator Michael W. Weiner, MD. The primary goal of ADNI has been to test whether serial magnetic resonance imaging, positron emission tomography, other biological markers, and clinical and neuropsychological assessment can be combined to measure the progression of mild cognitive impairment and early Alzheimer’s disease\footnote{For up-to-date information, see \url{www.adni-info.org}.}. The data acquired from the ADNI dataset included all subjects from ADNI 1, ADNI 2, ADNI 3, and ADNI GO that have paired MRI and FDG-PET. The J-ADNI was launched in 2007 as a public-private partnership, led by Principal Investigator Takeshi Iwatsubo, MD. The primary goal of J-ADNI has been to test whether serial magnetic resonance imaging (MRI), positron emission tomography (PET), other biological markers, and clinical and neuropsychological assessment can be combined to measure the progression of late mild cognitive impairment (MCI) and mild Alzheimer’s disease (AD) in the Japanese population. Our in-house dataset is a well-characterized, single-site clinical dataset containing healthy controls and subjects with two different types of dementia.
Overall, there are 626 subjects of healthy control (CN) and subjects without evidence for a neurodegenerative disease (ND), 742 subjects with mild cognitive impairment (MCI), 451 subjects with Alzheimer's disease (AD), and 70 subjects with frontotemporal lobar degeneration (FTLD) used in our study. Additional details on the study population are presented in Table~\ref{tab:demographics} and Extended Data Table~\ref{tab:mri_measures}.

\subsubsection{Data splitting}

We split the data into training, validation, and test sets, making sure that each subject contributed only one scan, which was acquired at the baseline visit. This ensured a strict separation of subjects across the sets.
To prevent biased results from confounding factors, we balanced the distribution of diagnosis, age, and gender across all sets. We used a data splitting method adapted from ClinicaDL~\cite{clinicadl,routier2021clinica}. 
This method first calculates a propensity score for each sample, which is the probability of the sample belonging to the training set, using a logistic regression model with known confounders. The balance of the split is then assessed by comparing the percentiles of the propensity score distributions across the three sets and measuring the maximum deviation~\cite{ho2007matching}. This process is repeated for 1,000 randomly generated partitions, and the partition with the lowest imbalance is chosen for the final split.
As a result, on the merged dataset including all three cohorts (ADNI~\cite{adni}, J-ADNI~\cite{jadni}, and in-house data), we have 1,467 samples for training, 173 samples for validation, and 212 samples for testing. 
For the in-house dataset, we have 196 samples for training, 65 samples for validation, and 62 samples held out for evaluation.

\subsection{Data processing}

\subsubsection{PET processing}

FDG-PET scans preprocessing steps include co-registration of the raw FDG-PET frames for correction of head motion, averaging of the co-registered frames, mapping the averaged image into a standard $160 \times 160 \times 96$  image grid with $1.5\times1.5\times1.5~mm^3$ voxel size, intensity normalization, and filtering of the normalized image with a scanner-specific kernel to produce an image with isotropic resolution of 8 mm full-width-at-half-maximum (FWHM). 
The scans were additionally processed using SPM12\footnote{https://www.fil.ion.ucl.ac.uk/spm/software/spm12}. 
First, the origin of all images was set to the anterior commissure region, which is required for the normalization function in SPM. Secondly, the scans are spatially normalized to the anatomical space of the Montreal Neurological Institute (MNI) with the SPM normalization tool, the MNI152 template (ICBM 2009c non-linear symmetric template), and 4-th degree B-splines to interpolate the transformed image to $1.5\times1.5\times1.5~mm^3$ voxel size. These steps were performed sequentially using a MATLAB script.
We further perform skull-stripping on all PET scans using Synthstrip~\cite{synthstrip}. All data are min–max rescaled to the image intensity values between 0 and 1 to increase the homogeneity of the data. 

\subsubsection{MRI processing}

MRI scans are first processed using Freesurfer~\cite{fischlFreeSurfer2012} to obtain subcortical segmentations for the standardization, including bias field correction with the N4ITK method, linear (affine) registration using the SyN algorithm from ANTs to register each image to the MNI space (ICBM 2009c non-linear symmetric template), and skull-stripping.
All MRI scans are finally registered to their corresponding PET scans. The rigid registration between MRI and PET is performed individually within each subject to align the modalities, without sharing registration parameters between different samples. All data are min–max rescaled to the image intensity values between 0 and 1 to increase the homogeneity of the data. The final image size for both modalities is $113 \times 137 \times 113$. To eliminate most blank backgrounds, we further center-crop all scans to $112 \times 112 \times 112$. All scans are then resized to $80 \times 80 \times 80$ as the input to the model.

\subsubsection{Auxiliary data processing}
\label{sec:clnical_data_process}

The auxiliary clinical data $\boldsymbol{c}$ includes 13 variables: age, gender, education, cognitive examination scores MMSE and ADAS-Cog-13, genetic risk factor ApoE4, and MRI brain segmentation volumes obtained by Freesurfer~\cite{fischlFreeSurfer2012}, including cerebrospinal fluid volume, the total grey matter volume, cortical white matter volume, left hippocampus volume, right hippocampus volume, left entorhinal thickness, and right entorhinal thickness. 
For the in-house auxiliary data, only age, gender, and the MRI brain segmentation volumes are available. We further standardized each group of auxiliary variables to a mean of 0 and a standard deviation of 1, before integrating them into the framework. Details on the distribution of these auxiliary variables across different disease categories in different datasets are presented in Table~\ref{tab:demographics} and Extended Data Table~\ref{tab:mri_measures}.

\subsubsection{PET visualization with 3D-SSP maps}
\label{sec:method_3dssp}

NEUROSTAT 3D-SSP maps~\cite{minoshima1995diagnostic} (Neurological Statistical Image Analysis Software 3D Stereotactic Surface) is a statistical quantitative brain mapping tool widely adopted in clinical settings. It is designed to investigate brain disorders and assist clinical diagnosis using PET and has contributed to identifying functional abnormalities in various brain disorders.
By comparing a patient's PET scan against a database of age-matched healthy controls, 3D-SSP produces Z-score maps that reliably quantify the statistical significance of regional metabolic deviations. 
The process involves spatially transforming trans-axial brain images to match a 3D reference brain from a stereotactic atlas, extracting peak cortical metabolic activity values, and projecting them onto a surface rendition of the brain. The resulting projections are statistically compared pixel-wise against a database of PET scans in age-matched controls, producing Z-score maps that highlight significant deviations~\cite{3d-ssp-explain}.
This tool provides a quantitative method for visualizing pathological consistency on PET. Thus, we additionally provided the 3D-SSP visualizations of simulated PET images for the evaluation.

\subsection{SiM2P translation framework}

\subsubsection{Overview}

Our goal is to generate simulated PET scans from corresponding MRIs and supplementary patient clinical information, assisting real-life clinical decision-making. To achieve this, we adapt the framework of Denoising Diffusion Bridge Models (DDBMs)~\cite{DDBM} for this 3D conditional translation. 
This framework generalizes diffusion models to handle paired distributions (MRI and PET data in our case) rather than using a fixed noise prior. In conventional diffusion models, one starts from random noise and learns to produce data. In contrast, diffusion bridges learn a direct transformation between two data distributions. This makes them an intuitive fit for cross-modality image translation, where we have corresponding MRI to PET pairs and wish to learn their mappings. It also eliminates the need for external guidance or projection steps to incorporate the MRI information into the generation, as in the prior work~\cite{Li2024pasta}.
While recent methods like flow matching~\cite{lipman2023flow} and rectified flow~\cite{liu2023flow} also learn continuous-time transport between distributions, DDBMs provide a more general framework for constructing a stochastic bridge with a tractable intermediate marginal distribution.
As existing diffusion bridge frameworks have primarily been developed for 2D natural images, we adapted this framework to 3D medical imaging using a volumetric diffusion Transformer as the backbone, and further conditioned it on the patient's auxiliary clinical data. Together, this design enables high-fidelity translation from MRI to PET. By operating directly between the MRI and PET domains, the diffusion bridge implicitly captures complex cross-modal relationships without requiring handcrafted loss terms or auxiliary constraints. Below, we detail the mathematical formulation of the diffusion bridge and its associated training objective in Section~\ref{sec:diffusion_bridge_formulation}, the backbone architecture in Section~\ref{sec:backbone_archi}, auxiliary data integration and missingness handling in Section~\ref{sec:clinical_data_integration_missingness}, and finally some alternative model designs in Section~\ref{sec:alternative_model_designs}.


\subsubsection{Diffusion bridge model formulation}
\label{sec:diffusion_bridge_formulation}

\textbf{Forward diffusion process: }
We define a continuous forward SDE (stochastic differential equation) that drives the state from the PET domain toward the MRI domain over time $t \in [0, T]$. Let $\mathbf{x}_0\in\mathbb{R}^{H \times W \times D} = \mathbf{x}$ be the PET scan and $\mathbf{x}_T\in\mathbb{R}^{H \times W \times D} = \mathbf{y}$ be its corresponding MRI scan as the target prior distribution, and $\mathbf{c}\in\mathbb{R}^{n_c}$ the auxiliary data vector. The diffusion process ${\mathbf{x}_t: 0 \le t \le T}$ satisfies an SDE of the form:
\begin{equation}
    d \mathbf{x}_t = \mathbf{f}(\mathbf{x}_t, t)dt + g(t)d\mathbf{w}_t, \ \mathbf{x}_0 = \mathbf{x}, 
\end{equation}
where $\mathbf{f}: \mathbb{R}^d \times [0, T] \rightarrow \mathbb{R}^d$ is vector-valued drift function, $g: [0, T] \rightarrow \mathbb{R}$ is a scalar-valued diffusion coefficient, and $\mathbf{w}_t$ is a standard Wiener process (Brownian motion). This defines a standard diffusion process from the PET input ($\mathbf{x}$). To ensure this process arrives at the target domain MRI ($\mathbf{y}$), we can apply Doob's $h$-transform~\cite{doob1984classical}, adding a guiding drift term $\mathbf{h}(\mathbf{x}_t,t,\mathbf{y},T)$. Here, $\mathbf{h}(\mathbf{x}_t,t,\mathbf{y},T) = \nabla_{\mathbf{x}_t} \log p(\mathbf{x}_T \mid \mathbf{x}_t) |_{\mathbf{x}_T = \mathbf{y}}$ is the gradient of the log transition kernel from current state $t$ to the endpoint $T$, generated by the original SDE. Intuitively, $\mathbf{h}$ drives the diffusion trajectory in the direction that makes reaching $\mathbf{y}$ at time $T$ more likely. The drift-adjusted forward SDE becomes:
\begin{equation}
    d \mathbf{x}_t = \mathbf{f}(\mathbf{x}_t, t)dt + g(t)^2 \mathbf{h}(\mathbf{x}_t,t,\mathbf{y},T) + g(t)d\mathbf{w}_t, \ \mathbf{x}_0 = \mathbf{x}, \ \mathbf{x}_T = \mathbf{y}.
\end{equation}
This process can also be called a diffusion bridge, which is a stochastic trajectory that starts from $\mathbf{x}$ and is conditioned to end at $\mathbf{y}$. The added drift term $g(t)^2 \mathbf{h}$ continuously steers the random Brownian motion toward the target $\mathbf{y}$. As a result, for every PET-MRI pair $(\mathbf{x}, \mathbf{y})$ in our training data, we can imagine an instance of such a diffusion bridge connecting them (Fig.~\ref{fig:overview}b). \\

\noindent\textbf{Reverse diffusion process: }
During the reverse diffusion process, we can generate a PET scan from a new MRI by solving the corresponding reverse SDE. Our time-reversed SDE can be written as:
\begin{equation}
    d\mathbf{x}_t = \left[ \mathbf{f}(\mathbf{x}_t, t) - g^2(t) \left( \mathbf{s}(\mathbf{x}_t, t, \mathbf{y}, T) - \mathbf{h}(\mathbf{x}_t, t, \mathbf{y}, T) \right) \right] dt + g(t) d\widehat{\mathbf{w}}_t, \ \mathbf{x}_T = \mathbf{y},
\end{equation}
in which $\mathbf{s}(\mathbf{x}_t,t,\mathbf{y},T) = \nabla_{\mathbf{x}_t} \log q(\mathbf{x}_t \mid \mathbf{x}_T) |_{\mathbf{x}_T = \mathbf{y}}$ is the true score of the intermediate state given the endpoint, and $\widehat{\mathbf{w}}_t$ is the reverse-time Brownian motion.
This reverse SDE runs backward from $t=T$ to $0$ and samples from $q(\mathbf{x}_t|\mathbf{x}_T)$, the conditional probability of earlier states given the final state. In practice, as the exact conditional distribution $q(\mathbf{x}_t|\mathbf{y})$ for real data is unknown, $\mathbf{s}(\mathbf{x}_t,t,\mathbf{y},T)$ is also unknown and needs to be approximated with a neural network. Thus, the training of the diffusion bridge process is to learn a score function $\mathbf{s}_\theta$ that approximates $\mathbf{s}(\mathbf{x}_t,t,\mathbf{y},T)$ in the above equation, by matching against a tractable quantity, as in the denoising score-matching~\cite{songscore2021}. Once learned, we can plug this estimate into the reverse SDE and simulate from $t=T$ to $0$, starting from an initial $\mathbf{x}_T$ drawn from the MRI prior. 
For the sampling process, we employed the higher-order hybrid solver from the DDBM framework~\cite{DDBM}. This solver is built upon a second-order Heun sampler, which discretizes the sampling steps into $t_0<t_1\dots <t_{N_\text{step}}$ with decreasing intervals~\cite{karras2022elucidating}. It then introduces an additional scheduled Euler-Maruyama step, following the backward SDE in between the higher-order ODE (ordinary differential equation) steps. This hybrid approach combines the speed of an ODE solver with the stochasticity of an SDE. This helps avoid potential averaged or blurry outputs, which is crucial for high-fidelity generation. We set the number of sampling steps $N_\text{step}$ to be 100 in our study. A detailed analysis of the trade-off between generation quality and sampling runtime with different $N_\text{step}$ values can be found in Section~\ref{sec:quality_wrt_runtime}.
\\

\noindent\textbf{Neural network parameterization: }
We leverage a 3D diffusion Transformer $F_{\theta}$ with parameter $\theta$ to model the score function. It takes as input the noisy volume $\mathbf{x}_t$ at time $t$, along with the patient's auxiliary data $\mathbf{c}$. 
We incorporate $\mathbf{c}$ by concatenating its embedding to the timestep embedding, driving the model towards predictions based on patient-specific factors. 
Following the pred-$\mathbf{x}$ parameterization in Elucidating Diffusion Models (EDM)~\cite{karras2022elucidating}, the final model output $D_{\theta}$ can be written as:
\begin{equation}
    D_\theta (\mathbf{x}_t,t, \mathbf{c})=c_{\text{skip}}(t)\mathbf{x}_t+c_{\text{out}}(t)F_\theta\big(c_{\text{in}}(t)\mathbf{x}_t, c_{\text{noise}}(t), \mathbf{c} \big).
\end{equation}
Let $a_t=\alpha_t/\alpha_T\ast \text{SNR}_T/\text{SNR}_t$, $b_t=\alpha_t(1-\text{SNR}_T/\text{SNR}_t)$, $c_t=\sigma_t^2(1-\text{SNR}_T/\text{SNR}_t)$, in which $\text{SNR}_t=\alpha_t^2/\sigma_t^2$ is the signal-to-noise ratio at time $t$, and $\alpha_t$, $\sigma_t$ are pre-defined signal and noise schedules. The scaling functions are derived to be:
\begin{align}
c_{\text{in}}(t) &= \frac{1}{\sqrt{a_t^2 \sigma_T^2 + b_t^2 \sigma_0^2 + 2 a_t b_t \sigma_{0T} + c_t}}, \\
c_{\text{out}}(t) &= \sqrt{a_t^2 (\sigma_T^2 \sigma_0^2 - \sigma_{0T}^2) + \sigma_0^2 c_t} \ast c_{\text{in}}(t), \\
c_{\text{skip}}(t) &= \left( b_t \sigma_0^2 + a_t \sigma_{0T} \right) \ast c_{\text{in}}^2(t), \\
c_{\text{noise}}(t) &= \frac{1}{4} \log(t) ,
\end{align}
where $\sigma_0^2$, $\sigma_T^2$, and $\sigma_{0T}$ are the variance of $\mathbf{x}_0$, $\mathbf{x}_T$, and the covariance of the two respectively. 
The signal and noise schedules $\alpha_t$, $\sigma_t$ are determined by the choice of the forward SDE. Two common Gaussian formulations are the variance exploding (VE) and variance preserving (VP) SDEs~\cite{songscore2021}. The VE SDE has zero drift and time-increasing diffusion, i.e., $f(\mathbf{x}_t,t)=0$ and $g(t)^2=d\sigma^2_t/dt$. Here, the signal scale is fixed, $\alpha_t=1$, and the noise schedule $\sigma_t=t$ is monotonically increasing with $t$. The VP schedule follows $f(\mathbf{x}_t,t)=(d\log\alpha_t/dt)\mathbf{x_t}$ and $g(t)^2=d\sigma^2_t/dt-2\sigma_t^2(d\log\alpha_t/dt)$. With a time-invariant drift $f(\mathbf{x}_t,t)=-0.5\beta_0\mathbf{x}_t$, we obtain $\alpha_t=\exp{(-0.5\beta_0t)}$ and $\sigma_t=\sqrt{1-\exp{(-\beta_0t)}}$. This leads to a bridge that has symmetric noise levels with respect to time~\cite{DDBM}. An experiment comparing these two bridge formulations can be found in Section~\ref{sec:alternative_model_designs}.
\\

\noindent\textbf{Training objective: }
We train $D_{\theta}$ by minimizing a weighted mean squared error between its prediction and the true $\mathbf{x}_0$ (clean PET), leading to the training objective:
\begin{equation}
    L(\theta) = \mathbb{E}_{\mathbf{x}_0, \mathbf{y}, t, \mathbf{c}} \big[ w(t) ||D_\theta(\mathbf{x}_t, t, \mathbf{c})-\mathbf{x}_0||^2 \big],
\end{equation}
where $w(t) = \frac{1}{c_{\text{out}}(t)^2}$ is a time-dependent weighting function. 
By matching the model's output to $\mathbf{x}_0$, we indirectly align $D_{\theta}$ with the diffusion bridge's score function. 
After training, the model can then be used for inference. Given a new MRI $\mathbf{y}$ and its auxiliary data $\mathbf{c}$, we can simulate the reverse diffusion starting from $\mathbf{x}_T = \mathbf{y}$ at time $t=T$, and integrate the learned reverse-time SDE to obtain a sample $\hat{\mathbf{x}_0}$ at time $t=0$, using the higher-order hybrid sampler~\cite{DDBM}.
The resulting $\hat{\mathbf{x}_0}$ is thus a simulated PET scan drawn from the model's learned conditional distribution. 
The training was done using the mini-batch strategy with the Adam optimizer~\cite{kingma2015adam}, with a learning rate of 0.0001 and no weight decay. The batch size is 1, and the training is done on a single NVIDIA H100 GPU. Each training iteration requires 1.86 seconds, with a peak GPU memory of 40.2 GB. We terminate the training at 180K iterations. During training, the model's performance was evaluated on the validation set at every 10K iterations, and the model with the highest performance was selected.

\subsubsection{Backbone architecture}
\label{sec:backbone_archi}

We implement a 3D diffusion Transformer as the backbone of our model. 
Given the 3D input $\mathbf{x}_t \in \mathbb{R}^{H \times W \times D}$, we first patchify it into a sequence of non-overlapping $h \times w \times d$ voxel patches. Let $n$ be the total number of patches, the patchification produces $\mathbf{x}_t^p \in \mathbb{R}^{n \times h \times w \times d}$, which we map to a sequence of patch embeddings $\mathbf{x}_t^e \in \mathbb{R}^{n \times f_e}$ via a learnable 3D convolution layer, with $f_e$ the dimension of the feature embeddings. Inspired by DiT-3D~\cite{mo2023dit3d}, we add frequency-based 3D sine-cosine positional embeddings to all patch embeddings $\mathbf{x}_t^e$ for better voxel structure locality across all axes. Next, the patch embeddings are processed by a sequence of $N$ Transformer blocks. Each block is conditioned on the diffusion timestep $t$ and auxiliary clinical data $\mathbf{c}$ using the adaptive layer normalization zero (adaLN-Zero) block~\cite{Peebles2022DiT}. 
We encode $t$ and $c$ independently using multilayer perceptrons (MLPs) into vector embeddings with length $f_e$, and these two embeddings are concatenated together as the input to the adaLN-Zero blocks. As shown in Extended Data Fig.~\ref{fig:method}, the adaLN-Zero predicts dimension-wise scale/shift parameters $\gamma$/$\beta$ and a residual scaling parameter $\alpha$, using MLP applied to the concatenation of $t$ and $\mathbf{c}$ embeddings. These parameters are applied within Transformer sub-layers, including global attention and fully connected layers, prior to the residual connections. The MLP is initialized to output a zero vector for all $\alpha$. In this way, the whole block is initialized as the identity function, a practice shown to benefit the overall training~\cite{Peebles2022DiT}.
After the final Transformer block, the patch embeddings go through a final adaptive layer norm and are linearly projected back to voxel patches. We then unpatchify them to reconstruct a dense 3D prediction that matches the input spatial resolution. In this study, we used a patch size of 4 ($h=w=d=4$), $N=28$ Transformer blocks, feature embedding dimension $f_e=1152$, and the number of attention heads 16. The total number of trainable parameters for the backbone model is 904M, requiring around 3.4 GB of memory.

The Transformer architecture is known to scale more effectively with increasing model size, training data, and compute resources than conventional convolutional networks~\cite{vaswani2017attention,dosovitskiy2020image}. This enhanced capacity allows diffusion Transformers to learn more complex, fine-grained relationships within the data~\cite{Peebles2022DiT}, which is critical for learning a robust structural-to-functional mapping from MRI to PET.
The adaLN-Zero pathway provides a clean and flexible approach to inject auxiliary data $\mathbf{c}$ as an additional condition, enabling the network to simulate metabolism patterns with patient-specific information.

\subsubsection{Auxiliary data integration and missingness handling}
\label{sec:clinical_data_integration_missingness}

We incorporate 13 variables (see Section~\ref{sec:clnical_data_process}) as auxiliary inputs to support the translation process and better reflect routine clinical practice, where such information typically accompanies MRI scans. To handle incomplete auxiliary clinical data, we append a binary missingness indicator to each variable, with 1 for present and 0 for missing, following the standard practice for tabular models~\cite{tabularselect,Poelsterl2021-daft}. This approach enables the network to make use of incomplete inputs while also learning patterns related to missingness. The final auxiliary input vector $\boldsymbol{c} \in \mathbb{R}^{n_c}$ thus comprises $n_c = 26$ features in total. We encode $\boldsymbol{c}$ to a fixed-width embedding using a three-layer MLP with SiLU activations between layers. This embedding is then concatenated with the timestep embedding and fed into the adaLN-Zero in each Transformer block. These low-dimensional priors provide complementary, patient-specific information to integrate with structural cues from the MRI. Empirically, this conditioning helps to improve the structural-to-functional mapping with enhanced accuracy of simulated PET (Section~\ref{sec:impact_clinical_data}).

\subsubsection{Analysis on the influence of individual auxiliary data}
\label{sec:method_auxiliary_influence}

We further conducted a sensitivity analysis to assess the influence of each auxiliary variable on the SiM2P simulation process. In this inference-time experiment, we isolated the effect of each of eleven variables by preserving its original value while setting all others to their respective dataset mean. 
These variables include age, gender, education level, ApoE4, MMSE, ADAS-Cog-13, as well as MRI-derived segmentation volumes from FreeSurfer, including cerebrospinal fluid (MRISeg-CSF), total grey matter volume (MRISeg-GM), cortical white matter volume (MRISeg-WM), left and right hippocampal volumes (MRISeg-Hippo), and left and right entorhinal thickness (MRISeg-Ent) (Extended Data Fig.~\ref{fig:clinical_data_influence}).

\subsubsection{Downstream automated classification task}
\label{sec:downstream_classification}

To further benchmark the diagnostic utility of the simulated PET, we performed a downstream automated classification task for the three-class differential diagnosis of dementia (CN-versus-AD-versus-bvFTD). 
We employed a 3D ResNet-18 model as the classifier. The classifier was trained and evaluated on our in-house dataset splits using various input modalities, including MRI, MRI with auxiliary clinical data (MRI+c), original PET, and simulated PET from SiM2P and competing diffusion model-based methods, PASTA and BBDM.

\subsubsection{Alternative model designs}
\label{sec:alternative_model_designs}

\paragraph{Alternative approaches for auxiliary data integration}

We investigated three representative approaches, concatenation, addition, and multiplication, for integrating the auxiliary data embeddings with the timestep embedding into the 3D diffusion Transformer. On the validation set, both concatenation and addition yielded comparable high performance, with an MAE of $0.0196\pm0.0054$ and $0.0191\pm0.0052$, PSNR of $28.62\pm1.83$ and $28.57\pm1.80$, respectively. Multiplication demonstrated a slightly lower performance, with an MAE of $0.0211\pm0.0057$ and a PSNR of $27.82\pm1.81$. We selected concatenation for the final model due to its inherent architectural flexibility, as it preserves the distinct information from both the auxiliary clinical variables and the diffusion timestep, allowing the model to learn their complex interactions.

\paragraph{Incorporating additional supervision from 3D-SSP maps}

NEUROSTAT 3D-SSP maps~\cite{minoshima1995diagnostic} provide a trusted, quantitative method for visualizing pathological consistency on PET, making it a natural candidate as an extra supervision signal to improve the clinical fidelity of simulated PET.
Thus, we explored incorporating 3D-SSP maps as an additional supervision signal in SiM2P. Specifically, we investigated two strategies. The first introduces a pretrained projection network that maps generated PET scans into 3D-SSP space and compares them against the corresponding real SSP maps during training. The second one aims to directly integrate 3D-SSP maps into the diffusion bridge, where the generation endpoint consists of both the PET volume and its associated 3D-SSP map, rather than the PET alone. An ablation study was conducted to evaluate the contribution of each approach.
However, both strategies led to similar performance but did not improve upon the baseline model, which was not given 3D-SSP map supervision. The two approaches yielded an MAE of $0.0219 \pm 0.0053$ and $0.0219 \pm 0.0056$, respectively, which is significantly higher than the baseline's $0.0196 \pm 0.0054$ ($p<0.0001$). While the first approach achieved an SSIM of $0.934 \pm 0.013$, comparable to the baseline's $0.939 \pm 0.012$, the overall performance did not improve. 
This suggests that our baseline model, trained with the full 3D PET volume alone, already implicitly learns the key hypometabolic patterns and clinical features that the 3D-SSP maps summarize. Explicitly adding this extra supervision signal appears to constrain the learning process, potentially causing a loss of fine-grained spatial information and reducing the model's overall generation fidelity.

\paragraph{Alternative diffusion bridge formulations}

We investigated implementing the diffusion bridge using two different score-based SDE formulations, variance exploding (VE) and variance preserving (VP) SDEs, and evaluated their effects on the generation quality. In both cases, the bridge yields analytic intermediate marginals enabling the denoising objective described earlier. Our results showed that the VP bridge significantly outperformed the VE bridge in our application, with the VE bridge yielding an MAE of $0.0274 \pm 0.0043$ compared to VP's $0.0196 \pm 0.0054$. In terms of SSIM, the VE bridge also led to a significant drop to $0.727 \pm 0.024$, compared to $0.939 \pm 0.012$ from the VP bridge ($p<0.0001$).

This finding aligns with the performance trends observed in the original paper for denoising diffusion models~\cite{DDBM}.
We hypothesize that this is a critical choice for medical image generation. The VP bridge's design, which maintains a bounded variance throughout the diffusion process, is inherently suited to preserving the fine-grained structural and anatomical fidelity of medical scans. In contrast, the unbounded noise of the VE bridge may degrade or obscure subtle, yet clinically critical, anatomical details and pathological patterns, which are essential for generating a reliable PET simulation. The more controlled noise schedule of the VP bridge, therefore, is crucial for maintaining the quantitative accuracy and clinical fidelity required for our application.

\subsubsection{Generation quality with respect to sampling runtime}
\label{sec:quality_wrt_runtime}

We conducted an experiment to explore the trade-off between SiM2P's generation quality and sampling runtime, determined by the number of sampling steps $N_\text{step}$. The results demonstrated that while generation quality increased with more steps, the improvements showed diminishing returns beyond a certain point (Extended Data Fig.~\ref{fig:generation_quality_runtime}).
As $N_\text{step}$ increased, we observed a steady improvement in all image quality metrics. The most substantial gains occurred from $N_\text{step}=10$ to $N_\text{step}=30$, where the mean MAE decreased from $0.0346$ to $0.0218$ and the mean SSIM increased dramatically from $0.483$ to $0.833$. Beyond this range, the improvement began to plateau. Increasing $N_\text{step}$ from 80 to 100 resulted in a fair improvement in MAE from $0.0196$ to $0.0192$ and SSIM from $0.932$ to $0.939$, while $N_\text{step}$ from 100 to 180 led to only a minimal change in MAE from $0.0192$ to $0.0191$, while the sampling runtime increased substantially from 2.3 minutes to 4.1 minutes per sample. The sampling runtime increased almost linearly with the sampling steps, from approximately 0.22 minutes per sample when $N_\text{step}=10$ to 4.1 minutes at $N_\text{step}=180$. Our selected value of $N_\text{step}=100$ represents an effective balance, providing near-optimal generation quality without incurring the disproportionate computational cost of higher sampling steps.

\subsection{Baselines implementation}

We benchmarked SiM2P against four state-of-the-art generative models, including the GAN-based Pix2Pix~\cite{pix2pix} and ResViT~\cite{resvit}, and the diffusion model-based BBDM~\cite{bbdm} and PASTA~\cite{Li2024pasta}. 
These methods were chosen to represent diverse approaches, from general image-to-image translation (Pix2Pix~\cite{pix2pix}, BBDM~\cite{bbdm}), to medical-specific frameworks (ResViT~\cite{resvit}, PASTA~\cite{Li2024pasta})
ResVit~\cite{resvit} is a GAN-based method that integrates ResNet and ViT as backbones, designed for medical image translation.
PASTA~\cite{Li2024pasta} leverages conditional diffusion models and is specifically targeting MRI to PET translation.
All methods were evaluated on the merged dataset from all three cohorts (ADNI~\cite{adni}, J-ADNI~\cite{jadni}, and in-house dataset). We further adopted the two best-performed methods, BBDM and PASTA, in the evaluation of the Local-Adapt framework.

\subsection{Clinical reader study}
\label{sec:method_clinical_reader_study}

\subsubsection{Study population}

The clinical reader study was conducted on a held-out test set of our in-house dataset. This subset consisted of 62 individuals, with a near-even distribution of cognitively normal (CN) controls (n = 22), patients with Alzheimer’s disease (AD, n = 19), and patients with behavioral-variant frontotemporal dementia (bvFTD, n = 21). 
This design helped ensure balanced representation of disease states. 
All AD and bvFTD patients were referred to TUM University Hospital (Munich, Germany)~\cite{inhouse_data} for PET/MRI evaluation due to suspected neurodegenerative conditions. Their diagnoses were confirmed by a consensus of at least two experienced psychiatrists, based on a combination of clinical examinations, neuropsychological tests, laboratory results, multimodal imaging, and, when available, cerebrospinal fluid biomarkers. The imaging workup included structural MRI, FDG-PET, and, in some cases, amyloid PET.
The AD diagnoses followed established criteria from either the NINCDS-ADRDA~\cite{mckhann1984clinical} or NIA-AA~\cite{albert2011diagnosis}. Patients with mild cognitive impairment (MCI) were labeled as “due to AD” only if AD-specific biomarkers were present. Diagnoses for bvFTD adhered to internationally recognized criteria~\cite{rascovsky2011sensitivity}. In contrast, the CN participants were primarily recruited through local newspaper advertisements, and they showed no signs of psychiatric or neurological symptoms and reported no subjective cognitive complaints~\cite{inhouse_data}. 
All cases included in the clinical reader study have paired T1-weighted MRI scans and FDG-PET, along with information on their age and gender. 
Imaging was performed on a 3-T Siemens Biograph mMR scanner (Siemens Healthineers AG) under standard resting conditions. The structural T1-weighted (magnetization-prepared rapid gradient-echo) MRIs were acquired with a 3-dimensional normal gradient-recalled sequence (repeat time, 2,300.0 ms; echo time, 2.98 ms; $9.0\degree$ flip angle) measuring 160 sagittal slices (field of view, 240 $\times$ 256 mm; pixel spacing, 1 mm; 256 $\times$ 240 scan matrix; slice thickness, 1.0 mm). The PET acquisition was conducted in parallel for 15 min, starting 30 min after an intravenous injection of an average of 198 MBq of tracer (range, 154–237 MBq). 

The data included in the study were collected in accordance with the latest version of the Declaration of Helsinki after the consent procedures had been approved by the local ethics committee of Technische Universität München. Written informed consent was obtained from all subjects.

\subsubsection{Study design}
\label{sec:study_design}

The clinical reader study aims to evaluate the diagnostic performance of simulated PET relative to MRI. 
As MRI and PET interpretations require distinct expertise in two different fields, neuroradiology and nuclear medicine, we involved four expert raters in the study. Two neuroradiologists were responsible for interpreting the MRI scans: MRI Rater 1 (MRI-R1) and MRI Rater 2 (MRI-R2). Two nuclear medicine physicians performed diagnoses based on the simulated PET scans: PET Rater 1 (PET-R1) and PET Rater 2 (PET-R2).
All raters are board-certified clinicians with more than five years of experience.
The neuroradiologists received MRI scans from each subject and their auxiliary data (age and gender), while the nuclear medicine physicians received the simulated PET and the same auxiliary data.
The raters were asked to make diagnoses using the provided data. All other clinical and biomarker data were withheld to reduce biases, and raters were also blinded to the diagnostic label distribution and to each other's assessments.
Each rater recorded two stages of diagnosis, including (i) an assessment of the presence or absence of any dementia disorders, along with their confidence level (low, moderate, or high), and (ii) if a dementia disorder was present, a classification of the pattern as either AD or bvFTD, with a corresponding confidence level (low, moderate, or high).

Given that diagnostic uncertainty directly impacts clinical utility, we then computed a confidence-weighted diagnostic accuracy, which gave more weight to high-confidence decisions while downweighting lower-confidence ones. 
Specifically, for each case $i$, the indicator of correctness ($y_i=1$ if correct, 0 otherwise) was multiplied by a weight $w_i$ reflecting the rater's confidence level (low = 1.0, moderate = 2.0, and high = 3.0). The weighted accuracy was then computed as the sum of these confidence-weighted predictions divided by the sum of all confidence weights, as $\sum w_iy_i/\sum w_i$. 
This metric provides a more clinically meaningful and informative measure of performance by accounting for the rater’s diagnostic certainty.

The simulated PET scans involved in the reader study were produced following our proposed local adaptation workflow, Local-Adapt, which first pre-trained SiM2P on the publicly available datasets (ADNI~\cite{adni} and J-ADNI~\cite{jadni}) to learn generalizable structure-to-function relationships, and then rapidly fine-tuned it on our local in-house dataset to capture site-specific imaging characteristics. A held-out test set of the in-house data was preserved for the evaluation in the clincal reader study.

\subsection{Statistical analysis}

We used one-way analysis of variance and the two-sided $\chi^2$ test for continuous and categorical variables, respectively, to assess the overall differences in the population characteristics between the diagnostic groups across the study cohorts. 
For diagnostic performance comparison in the clinical reader study, we performed a one-sided McNemar's test to identify statistically significant increases gained by simulated PET over MRI.
As the confidence-weighted diagnostic accuracy provided a continuous score instead of a binary correctness label, we used the one-sided Wilcoxon signed-rank test instead to assess the significance of performance differences between simulated PET and MRI for different diagnostic tasks.
In the downstream automated classification task for the differential diagnosis of dementia, we performed a one-sided McNemar's test to identify statistically significant differences between different modalities as input.
Pairwise statistical comparisons of generation performance with different image quality metrics were performed with the one-sided Wilcoxon signed-rank test. All statistical analyses were conducted at a significance level of 0.05.

\subsection{Performance metrics}

We measured mean absolute error (MAE), mean squared error (MSE), peak signal-to-noise ratio (PSNR), and structure similarity index (SSIM) for the quantitative evaluation of simulated PET compared to the real ones. Lower values of MAE, MSE, and higher values of PSNR, SSIM represent better generation quality. These quantitative metrics were initially calculated for the entire testing cohort, followed by a stratified analysis based on age, gender, and diagnosis subgroups.
For the diagnosis results of the clinical reader study, we reported the accuracy with 95\% Wilson confidence intervals, sensitivity, specificity, positive predictive value, negative predictive value, and confusion matrices for each rater. We further computed a confidence-weighted accuracy. It took into account each rater's per-case diagnostic confidence (low, moderate, or high), and gave more weight to high-confidence decisions while downweighting lower-confidence ones. For interrater reliability, we measured the Cohen’s kappa statistic $\kappa$~\cite{mchugh2012interrater} with 95\% Wilson confidence intervals. For the downstream automated classification task of the differential diagnosis of dementia, we generated both macro-averaged and micro-averaged ROC and PR curves for different input modalities. Per-class ROC and PR curves were also reported. From each ROC and PR curve, we further derived the area under the curve values (AUC and AP, respectively). Further, we computed micro- and macro-averaged AUC and AP values. The micro-averaged approach combines true positives, true negatives, false positives, and false negatives from all classes into a unified curve, providing a global performance metric. In contrast, the macro-averaged metric calculates individual ROC/PR curves for each class before computing their unweighted mean, disregarding potential class imbalances.

\subsection{Computational software and hardware}

Our software development used Python (version 3.10.14) and the models were developed using PyTorch (version 2.1.0). We used several other Python libraries to support data analysis, including pandas (version 2.2.1), scipy (version 1.12.0), torchvision (version 0.16.0), and scikit-learn (version 1.0.2), and for image processing including nibabel (version 5.2.1), monai (version 1.3.0), and torchio (version 0.19.6). NEUROSTAT software library was used to produce 3D-SSP maps from PET scans. Image processing involved SPM12 software with MATLAB (version R2021a), and Freesurfer (version 7.2). Training the model on a single NVIDIA H100 GPU on a shared computing cluster had an average runtime of 1.86 seconds per iteration, whereas the generation during inference took around two minutes per instance. Clinicians reviewed MRIs and simulated PET scans using 3D Slicer (version 4.11.2).

\section{Data availability}

Data from ADNI and J-ADNI are available from the LONI website at \url{https://ida.loni.usc.edu} upon registration and compliance with the data usage agreement. The in-house patient data from TUM University Hospital (Munich, Germany) are protected due to patient privacy.
We used the Montreal Neuroimaging Institute MNI152 template for image processing purposes, and the template can be downloaded at \url{http://www.bic.mni.mcgill.ca/ ServicesAtlases/ICBM152NLin2009}.

\section{Code availability}

Python scripts and help files are made available on GitHub (\url{https://github.com/Yiiitong/SiM2P}).

\section*{Disclosure}
Timo Grimmer has received consulting fees from Acumen, Advantage Ther, Alector, Anavex, Biogen, BMS; Cogthera, Eisai, Eli Lilly, Functional Neuromod., Grifols, Janssen, Neurimmune, Noselab, Novo Nordisk, Roche Diagnostics, and Roche Pharma; lecture fees from Anavex, Cogthera, Eisai, Eli Lilly, FEO, Grifols, Janssen, Novonordisk, Pfizer, Roche Pharma, Schwabe, and Synlab; and has received grants to his institution from Biogen, Eisai, and Eli Lilly.
No other potential conflict of interest relevant to this article was reported.

\section*{Acknowledgements} 
This work was supported by the German Research Foundation (DFG), the Munich Center for Machine Learning (MCML), and the DAAD programme Konrad Zuse Schools of Excellence in Artificial Intelligence, sponsored by the Federal Ministry of Research, Technology, and Space. 
We gratefully acknowledge the Leibniz Supercomputing Centre for providing the computing resources.

Data collection and sharing for this project was funded by the Alzheimer’s Disease Neuroimaging Initiative (ADNI) (National Institutes of Health Grant U01 AG024904) and DOD ADNI (Department of Defense award number W81XWH-12-2-0012). ADNI is funded by the National Institute on Aging, the National Institute of Biomedical Imaging and Bioengineering, and through generous contributions from the following: Alzheimer’s Association; Alzheimer’s Drug Discovery Foundation; Araclon Biotech; BioClinica, Inc.; Biogen Idec Inc.; BristolMyers Squibb Company; Eisai Inc.; Elan Pharmaceuticals, Inc.; Eli Lilly and Company; EuroImmun; F. Hoffmann-La Roche Ltd and its affiliated company Genentech, Inc.; Fujirebio; GE Healthcare; ; IXICO Ltd.; Janssen Alzheimer Immunotherapy Research \& Development, LLC.; Johnson \& Johnson Pharmaceutical Research \& Development LLC.; Medpace, Inc.; Merck \& Co., Inc.; Meso Scale Diagnostics, LLC.; NeuroRx Research; Neurotrack Technologies; Novartis Pharmaceuticals Corporation; Pfizer Inc.; Piramal Imaging; Servier; Synarc Inc.; and Takeda Pharmaceutical Company. The Canadian Institutes of Health Research is providing funds to support ADNI clinical sites in Canada. Private sector contributions are facilitated by the Foundation for the National Institutes of Health (\url{www.fnih.org}). The grantee organization is the Northern California Institute for Research and Education, and the study is coordinated by the Alzheimer’s Disease Cooperative Study at the University of California, San Diego. ADNI data are disseminated by the Laboratory for Neuro Imaging at the University of Southern California.

J-ADNI was supported by the following grants: Translational Research Promotion Project from the New Energy and Industrial Technology Development Organization of Japan; Research on Dementia, Health Labor Sciences Research Grant; Life Science Database Integration Project of Japan Science and Technology Agency; Research Association of Biotechnology (contributed by Astellas Pharma Inc., Bristol-Myers Squibb, Daiichi-Sankyo, Eisai, Eli Lilly and Company, Merck-Banyu, Mitsubishi Tanabe Pharma, Pfizer Inc., Shionogi \& Co., Ltd., Sumitomo Dainippon, and Takeda Pharmaceutical Company), Japan, and a grant from an anonymous Foundation.

\clearpage

\renewcommand{\figurename}{Extended Data Fig.}
\renewcommand{\thefigure}{\arabic{figure}}
\renewcommand{\theHfigure}{ED\arabic{figure}}
\renewcommand{\tablename}{Extended Data Table}
\renewcommand{\thetable}{\arabic{table}}
\renewcommand{\theHtable}{ED\arabic{table}}
\setcounter{figure}{0}
\setcounter{table}{0}

\begin{figure}[hp]
    \centering
    \includegraphics[width=1\linewidth]{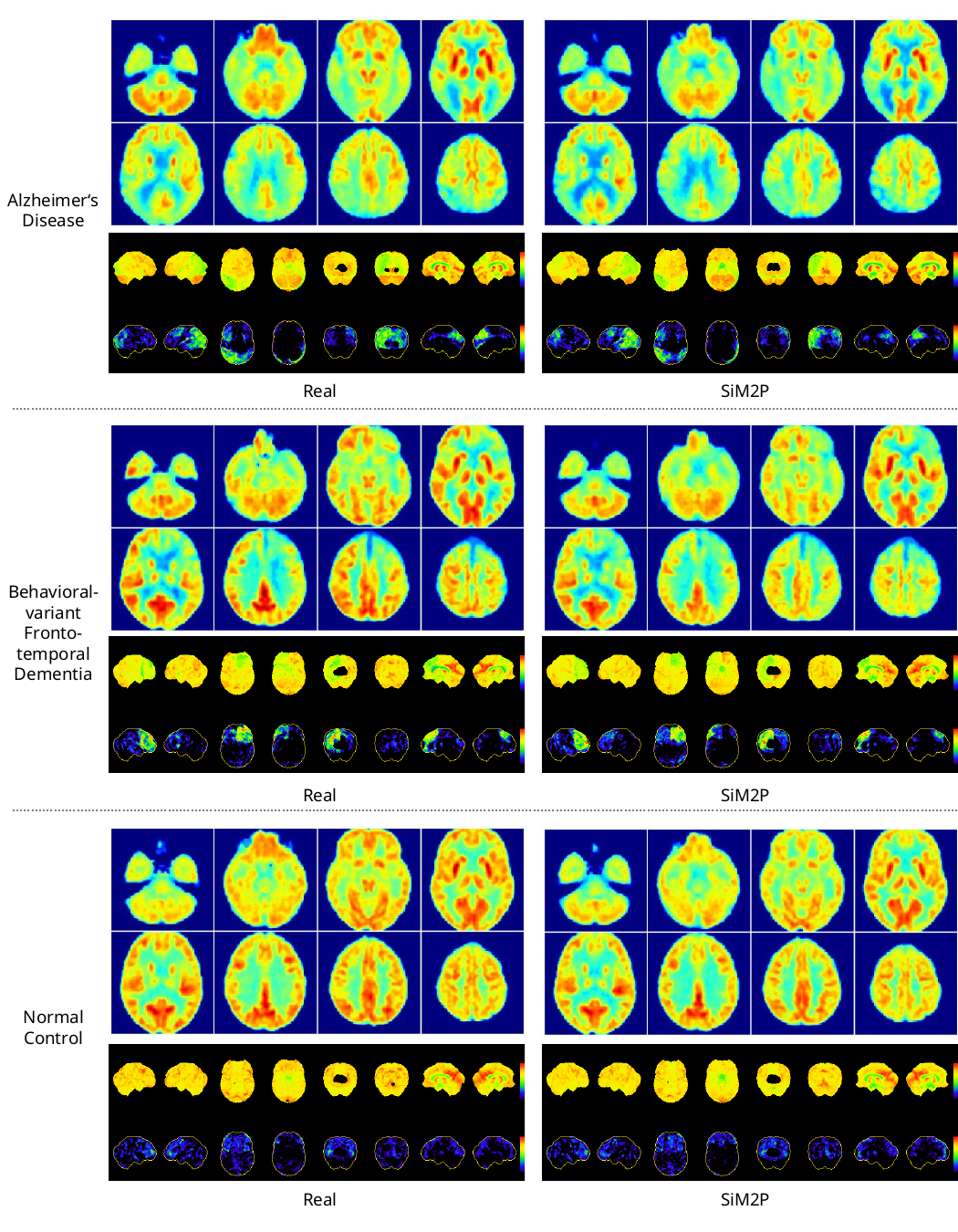}
    \caption{\textbf{Comparison of simulated PET from SiM2P and real PET.} For each subject (an Alzheimer's disease patient, a behavioral-variant frontotemporal dementia patient, and a normal control), we show eight evenly distributed axial slices of the PET scan. Below these slices, the corresponding 3D-SSP maps are displayed, where the first row shows a direct surface projection from different directions, and the second row provides a quantitative measure as a globally-normalized negative Z-score map.}
    \label{fig:qualitative_vis}
\end{figure}

\begin{figure}[hp]
    \centering
    \includegraphics[width=1\linewidth]{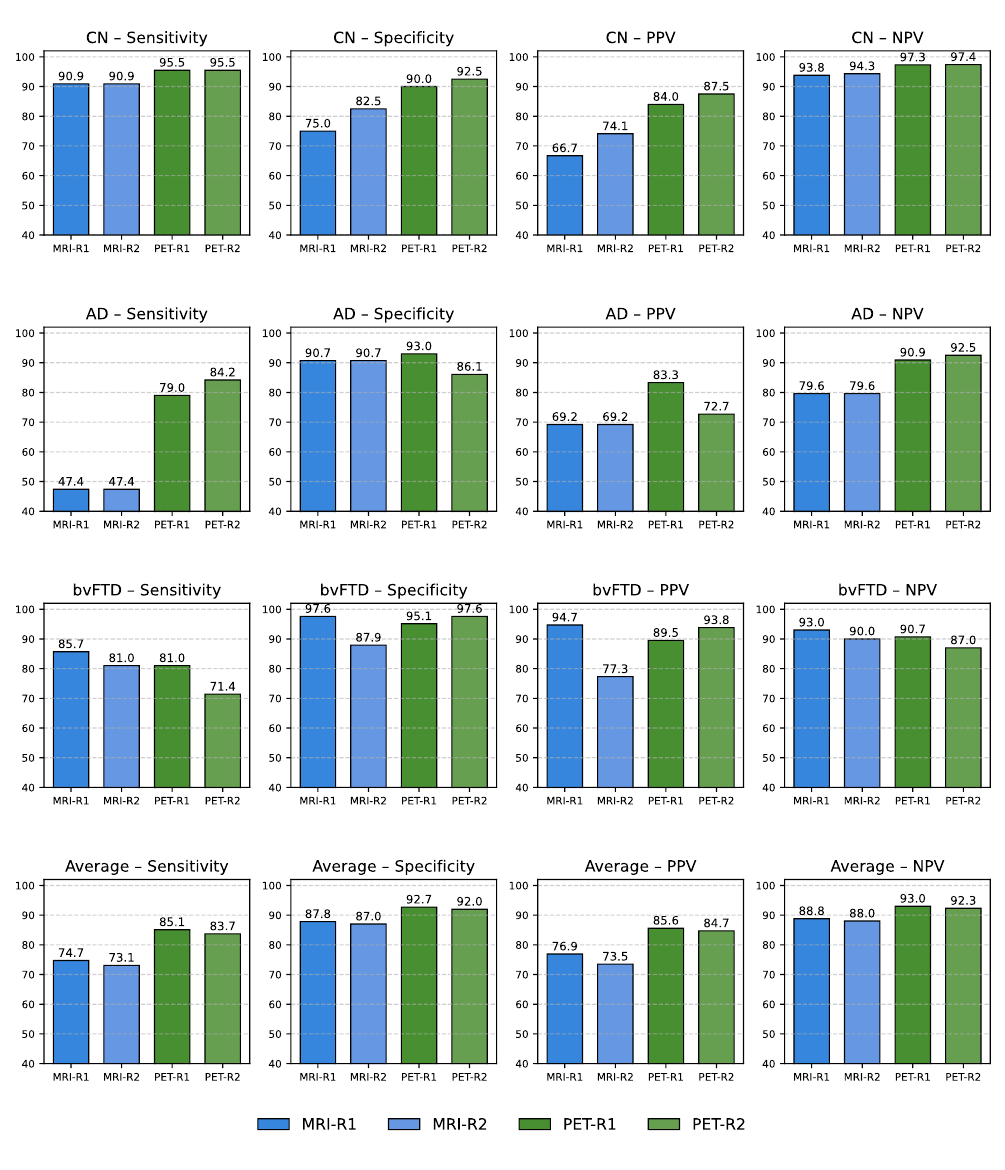}
    \caption{\textbf{Diagnostic performance of MRI (MRI-R1, MRI-R2) and simulated PET raters (PET-R1, PET-R2).} We demonstrate sensitivity, specificity, positive predictive value (PPV), and negative predictive value (NPV) for MRI and simulated PET raters R1/R2. The results are broken down by individual diagnostic categories and also include an overall average performance.}
    \label{fig:sen_spec}
\end{figure}

\begin{figure}[hp]
    \centering
    \includegraphics[width=1\linewidth]{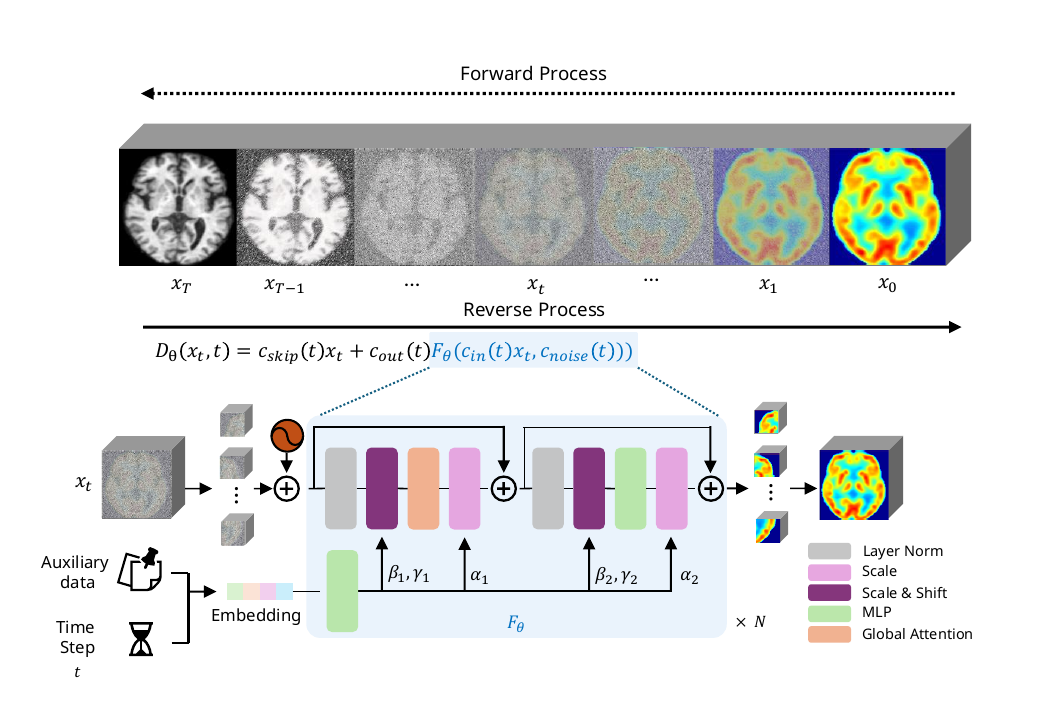}
    \caption{\textbf{Overview of the SiM2P architecture.} The SiM2P framework is designed to generate simulated PET scans from structural MRI and auxiliary clinical data via a 3D diffusion bridge process, modeling a stochastic mapping between the two modalities. The core of the model, denoted as $F_\theta$, is a 3D diffusion Transformer. It receives additional conditioning from auxiliary data, which is integrated by concatenating it with the timestep embedding as input to the adaptive layer norm zero (adaLN-Zero) blocks.}
    \label{fig:method}
\end{figure}

\begin{figure}[hp]
    \centering
    \includegraphics[width=1\linewidth]{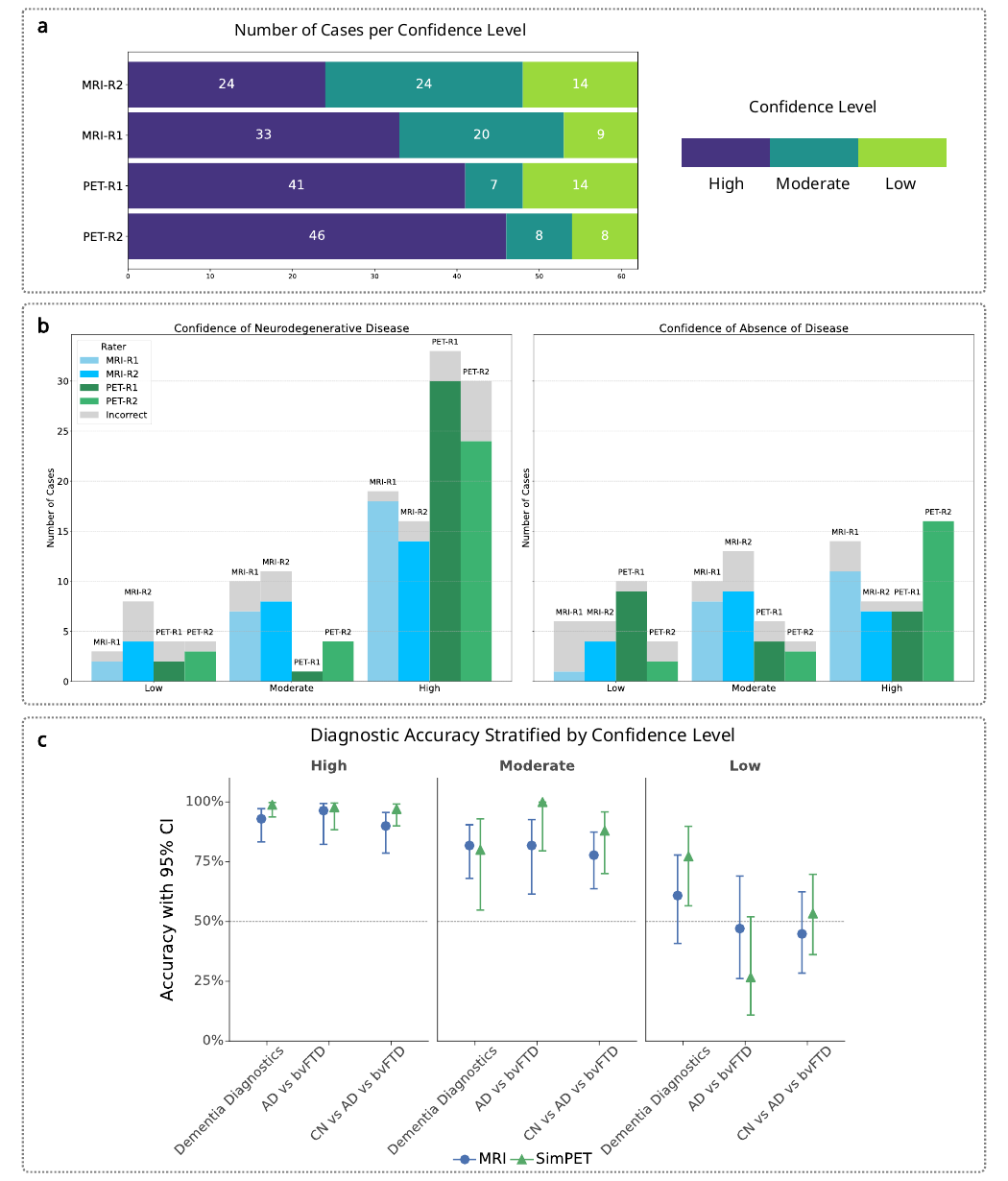}
    \caption{\textbf{Analysis of rater's per-case confidence level with respect to diagnostic correctness. a}, The distribution of diagnostic confidence levels (low, moderate, or high) assigned by each rater across all cases. \textbf{b}, The number of cases with correct or incorrect diagnosis for each rater stratified by their confidence levels. The results shown here are for the first stage of diagnosis, where the raters specify whether the case shows any evidence of dementia disorders or not. \textbf{c}, The diagnostic accuracy of MRI and simulated PET (SimPET) for all three tasks, with performance stratified by rater confidence levels.}
    \label{fig:confidence_analysis}
\end{figure}

\begin{figure}[hp]
    \centering
    \includegraphics[width=1\linewidth]{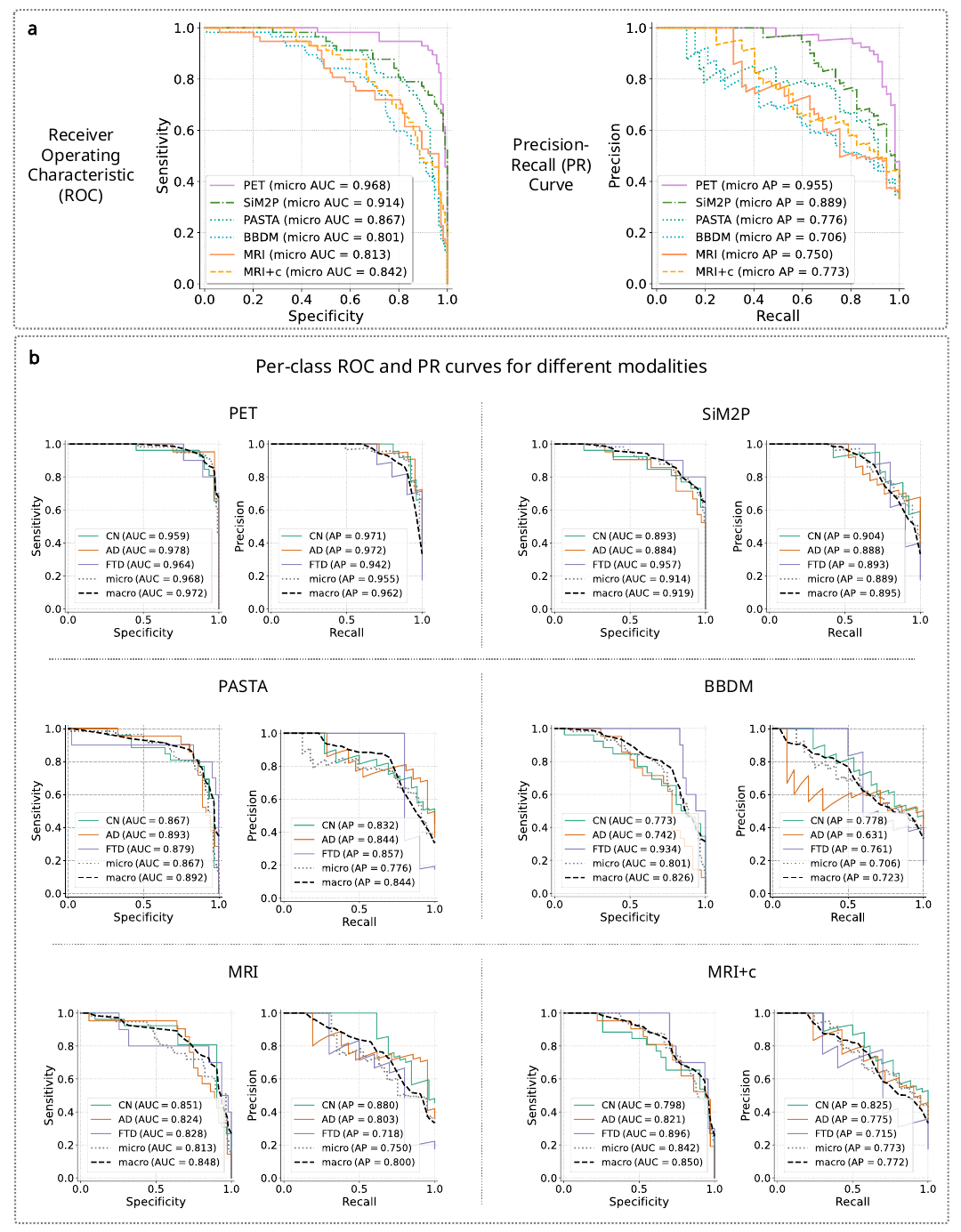}
    \caption{\textbf{Additional results from the automated differential diagnosis of dementia, using different modalities as input.} These inputs include real PET, simulated PET from SiM2P, PASTA, and BBDM, as well as MRI, and MRI with auxiliary clinical data (MRI+c). \textbf{a}, Micro-averaged Receiver Operating Characteristic (ROC) and Precision-Recall (PR) curves demonstrate overall model performance across all classes. \textbf{b}, Per-class ROC and PR curves show the performance for each diagnostic category across all types of inputs.}
    \label{fig:macro_micro_AP_AUC_plots}
\end{figure}

\begin{figure}[hp]
    \centering
    \includegraphics[width=0.92\linewidth]{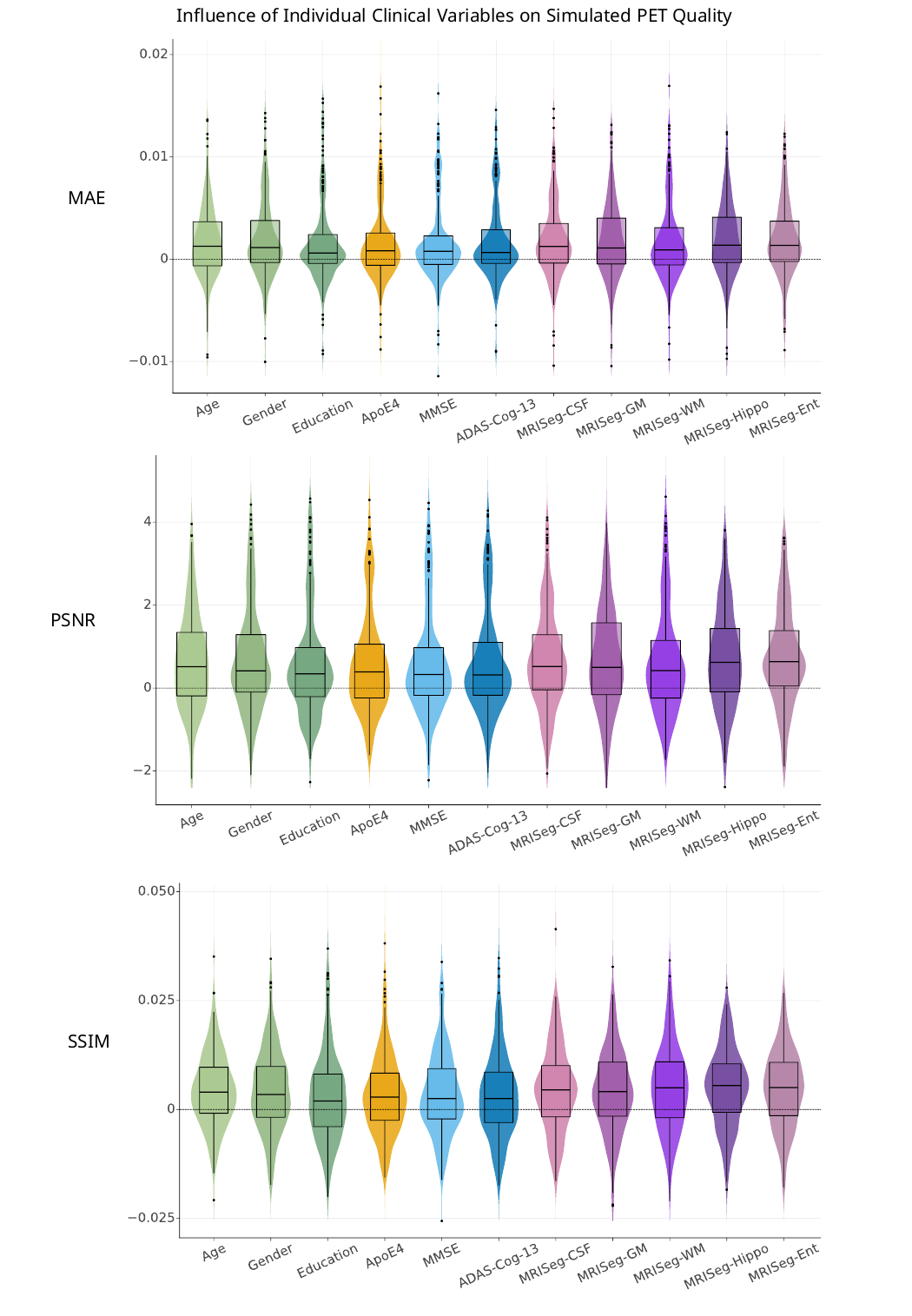}
    \caption{\textbf{Influence of individual auxiliary clinical variables on simulated PET quality.} We show the model's performance difference using a single clinical variable compared to the baseline configuration that uses all variables. A positive value on any metric indicates the magnitude of the performance drop. We measure the performance using MAE, PSNR, and SSIM. Boxplots show the median line and interquartile range (IQR) box, with whiskers extending to the most extreme values within 1.5 $\times$ IQR. The violin plots illustrate the distribution of results, with their width representing the density of data points at each value.}
    \label{fig:clinical_data_influence}
\end{figure}

\begin{figure}[hp]
    \centering
    \includegraphics[width=\linewidth]{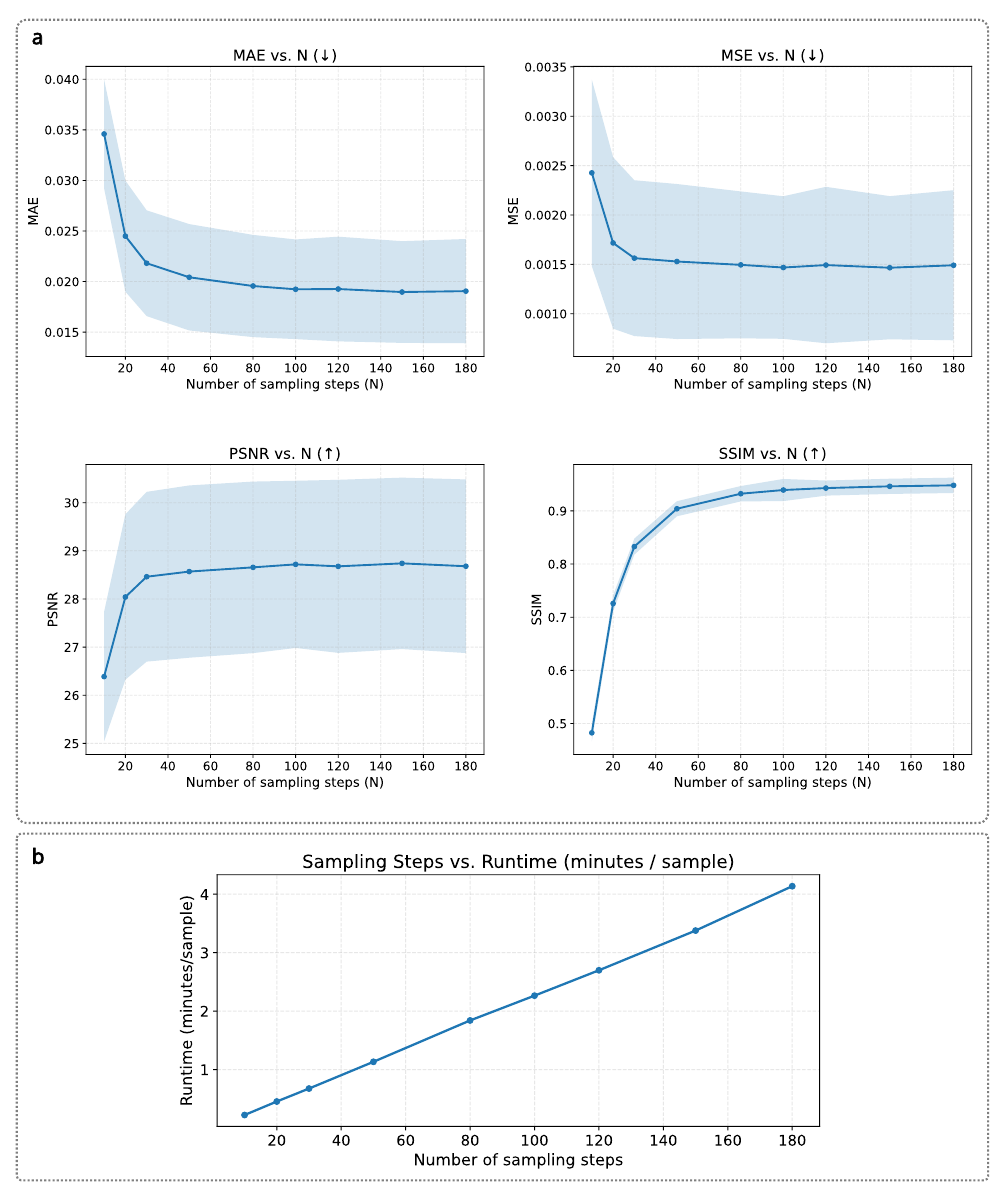}
    \caption{\textbf{Generation quality with respect to sampling runtime. a,} Mean (line) ± standard deviation (shaded band) of MAE, MSE, PSNR, and SSIM as a function of the number of sampling steps $N=10, 20, 30, 50, 80, 100, 120, 150, 180$, on the test set of the merged dataset. Quality improves monotonically with increasing $N$, with large gains from $N=10$ to $N=30$, and diminishing returns beyond $N\approx50$. Error bands denote cross-subject variability at each $N$. \textbf{b}, Sampling runtime (minutes) per sample with respect to the number of sampling steps.}
    \label{fig:generation_quality_runtime}
\end{figure}

\begin{table}[t]
\centering
\caption{\textbf{Quantitative evaluation of simulated PET quality across different methods.} We use four key metrics: mean absolute error (MAE$\downarrow$), mean squared error (MSE$\downarrow$), peak signal-to-noise ratio (PSNR$\uparrow$), and structure similarity index (SSIM$\uparrow$). Values are reported as the mean and standard deviation (mean$\pm$s.d) across the entire test set. For MAE and MSE, lower values indicate better performance, while for PSNR and SSIM, higher values are preferable. We show the results on the test set of our merged dataset (ADNI, J-ADNI, and in-house dataset), as well as on the test set of the in-house dataset after Local-Adapt.}
\label{tab:quantitative_results_comparison}
\renewcommand{\arraystretch}{1.5}
\begin{tabular}{cccccc}
\toprule
Data & Method & MAE$\downarrow$ & MSE$\downarrow$ & PSNR$\uparrow$ & SSIM$\uparrow$ \\
\midrule
 \multirow{6}{*}{\rotatebox{90}{Merged}} & Pix2Pix & 0.0896 $\pm$ 0.0047 & 0.0215 $\pm$ 0.0019 & 16.70 $\pm$ 0.3880 & 0.5330 $\pm$ 0.2748 \\
 & ResVit  & 0.0551 $\pm$ 0.0062 & 0.0140 $\pm$ 0.0027 & 18.61 $\pm$ 0.8382 & 0.7542 $\pm$ 0.2196 \\
 & BBDM    & 0.0285 $\pm$ 0.0062 & 0.0032 $\pm$ 0.0012 & 25.26 $\pm$ 1.4126 & 0.8992 $\pm$ 0.0203 \\
 & PASTA   & 0.0297 $\pm$ 0.0059 & 0.0036 $\pm$ 0.0012 & 24.58 $\pm$ 1.3128 & 0.9308 $\pm$ 0.0164 \\\cmidrule{2-6}
 & SiM2P & \multirow{2}{*}{\textbf{0.0192 $\pm$ 0.0049}} & \multirow{2}{*}{\textbf{0.0015 $\pm$ 0.0007}} & \multirow{2}{*}{\textbf{28.59 $\pm$ 1.7349}} & \multirow{2}{*}{\textbf{0.9393 $\pm$ 0.0139}} \\
 &  (Ours) &&&& \\
\midrule
\multirow{6}{*}{\rotatebox{90}{In-house}}
 & Pix2Pix & 0.1073 $\pm$ 0.0045 & 0.0271 $\pm$ 0.0021 & 15.68 $\pm$ 0.3178 & 0.4621 $\pm$ 0.0166 \\
 & ResVit  & 0.0491 $\pm$ 0.0055 & 0.0107 $\pm$ 0.0015 & 19.74 $\pm$ 0.5752 & 0.7687 $\pm$ 0.0230 \\
 & BBDM    & 0.0297 $\pm$ 0.0062 & 0.0034 $\pm$ 0.0011 & 24.93 $\pm$ 1.2777 & 0.8828 $\pm$ 0.0138 \\
 & PASTA   & 0.0321 $\pm$ 0.0067 & 0.0039 $\pm$ 0.0013 & 24.12 $\pm$ 1.3793 & 0.9204 $\pm$ 0.0180 \\\cmidrule{2-6}
 & SiM2P & \multirow{2}{*}{\textbf{0.0198 $\pm$ 0.0050}} & \multirow{2}{*}{\textbf{0.0017 $\pm$ 0.0008}} & \multirow{2}{*}{\textbf{27.90 $\pm$ 1.6127}} & \multirow{2}{*}{\textbf{0.9455 $\pm$ 0.0118}} \\
 &  (Ours) &&&& \\
\bottomrule
\end{tabular}
\end{table}

\begin{table}[hp]
\centering
\caption{Performance of SiM2P stratified based on age, gender, and diagnosis subgroups on the merged dataset.}
\renewcommand{\arraystretch}{1.8}
\begin{tabular}{cccccc}
\toprule
Demographics & Groups & MAE$\downarrow$ & MSE$\downarrow$ & PSNR$\uparrow$ & SSIM$\uparrow$ \\
\midrule
\multirow{4}{*}{Age} 
 & $<60$      & 0.0196 $\pm$ 0.0038 & 0.0015 $\pm$ 0.0005 & 28.33 $\pm$ 1.29 & 0.9339 $\pm$ 0.0130 \\
 & 60--70     & 0.0180 $\pm$ 0.0037 & 0.0013 $\pm$ 0.0006 & 29.05 $\pm$ 1.54 & 0.9414 $\pm$ 0.0152 \\
 & 70--80     & 0.0197 $\pm$ 0.0052 & 0.0015 $\pm$ 0.0008 & 28.43 $\pm$ 1.82 & 0.9391 $\pm$ 0.0141 \\
 & $>80$      & 0.0207 $\pm$ 0.0064 & 0.0017 $\pm$ 0.0010 & 28.19 $\pm$ 2.01 & 0.9380 $\pm$ 0.0097 \\
\midrule
\multirow{2}{*}{Gender} 
 & Male       & 0.0197 $\pm$ 0.0053 & 0.0015 $\pm$ 0.0008 & 28.46 $\pm$ 1.87 & 0.9379 $\pm$ 0.0157 \\
 & Female     & 0.0186 $\pm$ 0.0044 & 0.0014 $\pm$ 0.0006 & 28.74 $\pm$ 1.57 & 0.9409 $\pm$ 0.0112 \\
\midrule
\multirow{4}{*}{Diagnosis} 
 & CN         & 0.0185 $\pm$ 0.0041 & 0.0014 $\pm$ 0.0006 & 28.83 $\pm$ 1.55 & 0.9400 $\pm$ 0.0148 \\
 & MCI        & 0.0184 $\pm$ 0.0052 & 0.0013 $\pm$ 0.0008 & 29.04 $\pm$ 1.76 & 0.9438 $\pm$ 0.0092 \\
 & AD         & 0.0206 $\pm$ 0.0051 & 0.0017 $\pm$ 0.0008 & 27.97 $\pm$ 1.75 & 0.9341 $\pm$ 0.0156 \\
 & FTLD       & 0.0205 $\pm$ 0.0055 & 0.0016 $\pm$ 0.0008 & 28.11 $\pm$ 1.80 & 0.9373 $\pm$ 0.0133 \\
 \midrule
 \multicolumn{2}{c}{Total} &  \textbf{0.0192 $\pm$ 0.0049} & \textbf{0.0015 $\pm$ 0.0007} & \textbf{28.59 $\pm$ 1.73} & \textbf{0.9393 $\pm$ 0.0139} \\
\bottomrule
\end{tabular}
\end{table}

\begin{sidewaystable}[t]
\centering
\caption{MRI-derived volumetric and cortical thickness measures across three datasets, including cerebrospinal fluid volume (CSF), the total grey matter volume (Gray Matter Vol), cortical white matter volume (White Matter Vol), left hippocampus volume (Hippocampus (L)), right hippocampus volume (Hippocampus (R)), left entorhinal thickness (Entorhinal (L)), and right entorhinal thickness (Entorhinal (R)). Values are mean $\pm$ standard deviation. The $p$-value for each dataset indicates the statistical significance of intergroup differences per measure, computed with one-way ANOVA. (CN: healthy controls, AD: Alzheimer's disease, MCI: mild cognitive impairment, FTLD: frontotemporal lobar degeneration, ND: subjects without evidence for a neurodegenerative disease).}
\label{tab:mri_measures}
\renewcommand{\arraystretch}{1.8}
{\fontsize{7.5pt}{8.5pt}\selectfont
\begin{tabular}{lccccccc}
\toprule
Dataset (group) & CSF & Gray Matter Vol & White Matter Vol & Hippocampus (L) & Hippocampus (R) & Entorhinal (R) & Entorhinal (L) \\
\midrule
\textbf{ADNI} \\
CN [n=379]   & 1308.18 $\pm$ 317.05 & 553962.04 $\pm$ 49248.17 & 439687.33 $\pm$ 55265.17 & 3703.54 $\pm$ 507.23 & 3799.79 $\pm$ 529.38 & 3.49 $\pm$ 0.38 & 3.34 $\pm$ 0.32 \\
MCI [n=611]   & 1371.34 $\pm$ 363.47 & 550167.04 $\pm$ 52874.40 & 444981.59 $\pm$ 56792.36 & 3398.52 $\pm$ 624.71 & 3502.82 $\pm$ 615.37 & 3.31 $\pm$ 0.54 & 3.16 $\pm$ 0.50 \\
AD [n=257]   & 1504.75 $\pm$ 453.05 & 516933.17 $\pm$ 54817.95 & 429787.36 $\pm$ 63456.61 & 2796.31 $\pm$ 550.64 & 2922.55 $\pm$ 590.67 & 2.78 $\pm$ 0.55 & 2.65 $\pm$ 0.51 \\
$p$-value   & 4.91$\times 10^{-10}$ & 7.45$\times 10^{-20}$ & 1.93$\times 10^{-3}$ & 2.78$\times 10^{-73}$ & 3.49$\times 10^{-67}$ & 7.55$\times 10^{-63}$ & 7.89$\times 10^{-70}$ \\
\midrule
\textbf{J-ADNI} \\
CN [n=104]   & 1379.41 $\pm$ 415.93 & 528759.42 $\pm$ 45583.38 & 468728.24 $\pm$ 66153.44 & 3731.61 $\pm$ 515.64 & 3801.83 $\pm$ 541.13 & 3.57 $\pm$ 0.39 & 3.39 $\pm$ 0.34 \\
MCI [n=131]   & 1491.64 $\pm$ 471.61 & 500520.01 $\pm$ 46802.50 & 440698.65 $\pm$ 56597.60 & 2994.99 $\pm$ 579.53 & 3063.61 $\pm$ 604.95 & 3.07 $\pm$ 0.54 & 2.88 $\pm$ 0.49 \\
AD [n=84]   & 1561.95 $\pm$ 412.01 & 478350.80 $\pm$ 41470.11 & 436711.75 $\pm$ 59317.17 & 2648.84 $\pm$ 488.32 & 2659.51 $\pm$ 512.25 & 2.75 $\pm$ 0.56 & 2.61 $\pm$ 0.42 \\
$p$-value   & 2.04$\times 10^{-3}$ & 1.22$\times 10^{-18}$ & 4.51$\times 10^{-6}$ & 1.33$\times 10^{-52}$ & 1.84$\times 10^{-52}$ & 1.99$\times 10^{-35}$ & 7.76$\times 10^{-44}$ \\
\midrule
\textbf{In-house} \\
ND [n=143]   & 1243.21 $\pm$ 387.37 & 579154.99 $\pm$ 59578.73 & 425079.49 $\pm$ 65488.87 & 3753.01 $\pm$ 596.37 & 3865.57 $\pm$ 583.35 & 3.14 $\pm$ 0.36 & 3.00 $\pm$ 0.34 \\
AD [n=110]  & 1468.09 $\pm$ 528.13 & 542996.49 $\pm$ 52983.37 & 403678.20 $\pm$ 56971.95 & 3315.55 $\pm$ 539.39 & 3445.26 $\pm$ 540.94 & 3.01 $\pm$ 0.39 & 2.86 $\pm$ 0.41 \\
FTLD [n=70]  & 1487.83 $\pm$ 400.65 & 519333.68 $\pm$ 55307.27 & 380247.21 $\pm$ 49705.65 & 3126.89 $\pm$ 590.53 & 3167.09 $\pm$ 729.37 & 2.69 $\pm$ 0.68 & 2.60 $\pm$ 0.59 \\
$p$-value & 1.16$\times 10^{-5}$ & 8.91$\times 10^{-14}$ & 2.24$\times 10^{-7}$ & 3.91$\times 10^{-15}$ & 1.26$\times 10^{-15}$ & 4.48$\times 10^{-11}$ & 1.95$\times 10^{-10}$ \\
\bottomrule
\end{tabular}
}
\end{sidewaystable}


\clearpage
\printbibliography

\end{document}